\documentclass[11pt]{report}
\usepackage{makecell}
\usepackage{bm}
\usepackage{placeins}
\usepackage{longtable}
\usepackage{upennstyle} 
\usepackage{tcolorbox}
\tcbuselibrary{listings,breakable}
\usepackage[nonamebreak,round]{natbib}
\newcommand{\methodname}{Asset2Sim}
\title{AUTOMATICALLY IMPROVING SIMULATION PHYSICS FOR ARTICULATED OBJECTS}
\author{Anh Quan Pham}
\gradgroup{Robotics} 
\date{2026} 
\supervisor{Dinesh Jayaraman}
\supervisortitle{Assistant Professor, Department of Computer and Information Science}
\gradchair{M. Ani Hsieh, Associate Professor, Department of Mechanical Engineering and Applied Mechanics}

\authorlegal{Anh Quan Pham} 

\dedication{To my parents, for their unconditional support and trust in me.} 

\acknowledgement{I would first like to thank my advisor, Professor Dinesh Jayaraman, for his patience and dedication in mentoring me throughout my time at Penn. Working under Professor Jayaraman's guidance has taught me how to think more deeply about robotics research, approach impactful problems, develop a research philosophy, and reason through the unknown.

I would also like to thank my collaborators and friends in the PennPAL Lab for creating an incredibly supportive and intellectually stimulating environment. Their feedback and discussions were consistently constructive, and I am grateful for the camaraderie and encouragement, especially during long nights leading up to submission deadlines. I am especially grateful to Junyao Shi for his mentorship and guidance on both my research and career development.

Additionally, I would like to thank Marcel Hussing for giving me my first research opportunity at Penn when few others did. He taught me how to conduct rigorous machine learning research, to be patient with progress, and trusted me to take ownership of projects end-to-end independently. That experience has had a lasting influence on how I approach challenging problems and carry research efforts forward.

Parts of this thesis are based on collaborative work that has resulted in a co-authored publication currently under review. I am grateful to my co-authors—Sagnik Anupam, Luyang Hu, Kaitian Chao, George Jiayuan Gao, Tianyou Wang, and Junyao Shi—for their contributions and insights throughout this process, as well as to my co-advisors, Professors Osbert Bastani and Dinesh Jayaraman. A detailed breakdown of individual contributions is provided in Appendix~\ref{app:personal-contributions}.

Finally, I would like to thank my parents. Growing up in a small town, we did not have the means to travel or study in places with many opportunities, and going abroad was far beyond anything I could have imagined. Despite this, they found their own ways to bring the world to me. They taught me that being the first to pursue difficult paths comes with no predefined limits, and that belief continues to push me to go further every day.} 

\abstract{Simulation is a central tool for scalable robot learning, but its effectiveness depends on the quality of object assets. While modern 3D datasets provide rich geometric and kinematic representations, they typically lack the physical properties required for stable and realistic interaction, requiring significant manual effort to construct simulation-ready articulated objects.

In this thesis, we introduce \textit{interaction-readiness}, which characterizes whether an object can be reliably simulated under manipulation. We propose a quantitative evaluation framework that decomposes interaction-readiness into measurable components, enabling systematic analysis of object quality and revealing failure modes not captured by conventional evaluation.

We further present a multi-modal, simulator-in-the-loop approach for generating interaction-ready articulated objects from incomplete 3D assets. The method integrates geometric, visual, and semantic information to infer physical properties and refines them through iterative simulator feedback to improve physical consistency.

Experiments across diverse articulated objects and manipulation tasks show that object quality directly impacts simulation stability, interaction behavior, and policy performance. Objects refined by our method exhibit more stable and realistic dynamics, enabling more reliable downstream learning and evaluation.

Overall, this thesis demonstrates the importance of physical realism for articulated objects in simulation and introduces a practical multi-modal refinement approach, guided by simulator feedback, for constructing such objects at scale.} 

\begin{document}
\maketitle 
\setcounter{page}{2}

\makecopyright 

\makededication 

\makeacknowledgement 

\makeabstract
\tableofcontents

\clearpage \phantomsection \addcontentsline{toc}{chapter}{LIST OF TABLES} \begin{singlespacing} \listoftables \end{singlespacing}

\clearpage \phantomsection \addcontentsline{toc}{chapter}{LIST OF ILLUSTRATIONS} \begin{singlespacing} \listoffigures \end{singlespacing}


\begin{mainf} 
\chapter{INTRODUCTION}
\label{ch:introduction}

Modern robot learning increasingly depends on scale and diversity of data and training, particularly for manipulation tasks that are sensitive to variation in visual appearance and external physical properties. Unlike domains where structure is internal or can be abstracted away, robot manipulation is inherently grounded in the interactions with the physical world. Objects differ in geometry, material composition, mass distribution, and contact dynamics, and these variations directly affect the outcome of interactions. As a result, manipulation policies require diverse, physically plausible experiences to generalize robustly to real-world environments. Simulation has therefore emerged as a central pathway toward scalable robot learning.

Driven by this need for scalable and diverse interaction data, advances in large-scale 3D datasets and generation pipelines have enabled applications ranging from text-to-3D synthesis~\citep{poole2022dreamfusion} to robotics simulation and scalable environment construction, supported by platforms such as SAPIEN~\citep{Xiang_2020_SAPIEN}. Benchmarks such as RoboLab~\citep{yang2026robolab}, RoboCasa365 \citep{robocasa365}, and LIBERO~\citep{liu2023libero} further illustrate the use of simulation as a practical proxy for evaluating generalist robot policies. In parallel, automated environment generation systems such as RoboGen~\citep{wang2023robogen} and GenSim2~\citep{huagensim2} highlight the possibility of scaling simulated environments with minimal human intervention. Together, these developments point toward a trajectory in which large-scale simulation plays a central role in training and evaluating robot learning systems.

Despite this progress, a key limitation remains. The ability to scale simulated environments has outpaced the ability to construct interaction-ready objects, particularly articulated objects that support nontrivial manipulation. This limitation is evident in existing systems. Large-scale benchmarks such as RoboLab \citep{yang2026robolab} include many tasks but largely avoid articulated interactions. RoboCasa365 \citep{robocasa365} supports articulated objects such as cabinets and drawers, but doing so required substantial manual engineering effort and human expertise. Similarly, automated systems such as RoboGen \citep{wang2023robogen} and GenSim2~\citep{huagensim2} rely on a limited set of carefully curated articulated assets. In contrast, rigid objects can be scaled more easily due to their simpler physical structure. As a result, articulated object interaction, which is ubiquitous in real-world environments, remains underrepresented in simulation.

This gap reflects a fundamental issue in how objects are represented. Geometry and visual appearance are often sufficient for rendering and recognition, and modern datasets capture these properties at scale. However, for interaction, an object should be defined not only by its appearance but also by its physical behavior under contact and actuation. This includes properties such as mass, inertia, friction, damping, and joint dynamics. We refer to objects endowed with these properties as interaction-ready objects, meaning that they can be stably and realistically simulated under manipulation. We treat interaction-readiness as an operational construct for evaluating simulation usability, rather than a fundamental physical property. It is therefore instantiated as a composite of measurable criteria capturing different failure modes in simulation.

Constructing interaction-ready articulated objects remains challenging. Prior work on articulated object modeling has focused primarily on recovering geometric structure and kinematic relationships, for example through methods such as Articulate-Anything, Articulate-Anymesh, and URDF-Anything~\citep{le2024articulate, qiu2025articulate, li2025urdf}. While these approaches are effective at identifying links and joints, they do not address the problem of assigning physically consistent parameters required for stable simulation. Recent advances in vision-language models (VLMs) have shown promise in reasoning about material properties and physical attributes from multi-modal inputs~\citep{luo2023scalable, kabra2023leveraging}. However, these models operate as predictors and do not guarantee that their outputs yield physically valid or stable behavior when deployed in a simulator.

We identify two primary challenges. First, there is no widely adopted definition of interaction-readiness that can be evaluated quantitatively and at scale. Existing approaches often rely on downstream task success or qualitative inspection, which makes it difficult to isolate errors arising from object modeling. Second, generating physically consistent object models requires integrating geometry, appearance, and semantics with constraints imposed by physics. Inferring such properties from observations is inherently ambiguous, particularly for articulated objects, and purely predictive approaches may produce parameters that are plausible but physically inconsistent.

This thesis addresses these challenges through two contributions.

First, we introduce a quantitative evaluation framework for interaction-readiness. The proposed framework decomposes interaction-readiness into several measurable components, spanning physical stability, semantic alignment, behavioral fidelity, and perceptual realism. This decomposition allows object quality to be evaluated independently of downstream policies and enables systematic comparison across methods.

Second, we present a multi-modal, simulator-in-the-loop method for generating interaction-ready articulated objects. Rather than treating physical property estimation as a static prediction problem, this work reframes it as a simulator-grounded refinement process that iteratively enforces physical consistency. The approach uses a vision-language model to infer initial physical properties from geometric, visual, and semantic inputs, and refines them through closed-loop interaction with a physics simulator. This procedure corrects inconsistent parameters and improves object realism, enabling scalable refinement of simulation-ready articulated assets in a matter of seconds.

The contributions of this thesis are as follows:

\begin{itemize}

\item In Chapter~\ref{ch:related_work}, we provide a systematic analysis of prior work on articulated object modeling, physical property estimation, and simulation-based robot learning. We identify a key limitation in existing pipelines: while geometric and kinematic structures can be inferred at scale, current approaches lack mechanisms for assigning and validating physically consistent properties required for realistic interaction.

\item In Chapter~\ref{ch:evaluation-protocol}, we introduce a quantitative evaluation framework for interaction-ready objects. We operationalize interaction-readiness as a measurable construct and develop metrics for assessing physical stability, semantic alignment, behavioral fidelity, and qualitative realism, enabling principled comparison of object assets beyond downstream task success.

\item In Chapter~\ref{ch:method}, we present a multi-modal, simulator-in-the-loop approach for creating interaction-ready articulated objects from incomplete 3D assets. The method integrates geometric, visual, and semantic information to infer physical properties, and refines these estimates through iterative feedback from a physics simulator. This design addresses key limitations of model-only approaches by incorporating feedback to improve physical consistency and stability.

\item In Chapter~\ref{ch:experiments}, we conduct extensive evaluation across a range of articulated objects and manipulation tasks. Using the proposed evaluation framework, we demonstrate that our method produces more physically stable and behaviorally realistic assets, and that improvements in object quality directly impact downstream policy performance and perceived realism. We also analyze scalability and common failure modes in large-scale object generation.

\item In Chapter~\ref{ch:conclusion}, we discuss the implications of this work for scalable simulation-based robot learning and outline open challenges in constructing physically realistic and interaction-ready environments.

\end{itemize}

\chapter{\MakeUppercase{Related Work}}
\label{ch:related_work} 
 
\section{3D Asset Repositories and the Geometry-Physics Gap}
 
Large-scale 3D object datasets such as ShapeNet \citep{chang2015shapenetinformationrich3dmodel} and PartNet \citep{Mo_2019_CVPR} initially provided richly annotated models for use in environment creation sourced from the web. These repositories were groundbreaking in scale: ShapeNet contains over 51,000 models across 55 object categories, while PartNet provides hierarchical part decompositions for more than 26,000 objects. These datasets enabled downstream applications ranging from 3D shape analysis and recognition to texture synthesis and scene generation.
 
PartNet-Mobility \citep{Xiang_2020_SAPIEN} and GAPartNet \citep{geng2022gapartnet} extended those models with richer semantic and kinematic annotations. PartNet-Mobility \citep{Xiang_2020_SAPIEN}, in particular, was explicitly designed for robotics simulation: each model includes fine-grained part annotations and kinematic structure (which links are fixed, which move, joint types, and rotation axes). This advancement made it possible to programmatically construct simulation-ready articulated objects at scale. Benchmarks like SAPIEN \citep{Xiang_2020_SAPIEN} were built directly on this foundation.
 
However, these repositories deliberately leave masses, friction coefficients, damping values, and other physical properties largely unspecified or untuned for downstream tasks. Physical properties vary drastically based on material composition, manufacturing process, wear and aging, and intended use context. A wooden cabinet drawer behaves very differently from a plastic one, and a worn hinge has different damping than a new one. Inferring these properties from static 3D geometry alone is fundamentally ambiguous: it is impossible to determine an object's mass from its shape without knowing its material and whether it is solid or hollow. As a result, PartNet-Mobility \citep{Xiang_2020_SAPIEN} and similar datasets defer this problem to downstream users.
 
The practical consequence is significant. When researchers and practitioners export a kinematic asset from PartNet-Mobility \citep{Xiang_2020_SAPIEN} and loads it into a physics simulator like MuJoCo or Isaac Sim \citep{todorov2012mujoco, NVIDIA_Isaac_Sim}, the simulator faces a choice: either (1) apply conservative default values (unit mass per link, high friction ~0.5, zero damping), which produce physically implausible dynamics where objects stick to surfaces unrealistically, joints oscillate uncontrollably, and collisions are stiff, or (2) require the user to manually tune dozens of parameters until the behavior matches real-world intuition. Option (1) produces broken simulations; option (2) requires domain expertise and extensive trial-and-error.
 
Recognizing this bottleneck, newer benchmarks have invested heavily in manual curation. ArtVIP \citep{jin2025artviparticulateddigitalassets} provides a curated set of articulated objects with carefully specified mass distributions, center-of-mass offsets, joint stiffness, damping, and friction values. GenSim2 \citep{huagensim2}, which is designed to automatically generate robot manipulation environments for data collection, still relies on human annotators to validate and refine physical properties for each object before downstream policy learning. These benchmarks achieve high-quality learning environments, but at substantial cost: the ArtVIP dataset \citep{jin2025artviparticulateddigitalassets} required substantial expertise from professional 3D modelers to curate, and GenSim2 \citep{huagensim2} required significant human efforts for a relatively small set of objects. This manual bottleneck makes scaling to thousands of object categories impractical.
 
Efforts to automate property prediction have made limited progress for articulated systems. ObjectFolder 2.0 \citep{gao2022objectfolder} explored learning-based prediction of deformation and material properties for rigid objects, achieving promising results by training on large-scale observations of real objects. However, extending this to articulated objects faces two critical challenges: (1) obtaining ground-truth labels requires either physical measurement (mass, inertia tensors, friction coefficients via specialized laboratory equipment) or interactive robotic experiments (inferring properties from contact dynamics), both of which are expensive and difficult to parallelize, and (2) articulated objects introduce additional complexity---incorrect mass distribution can cause self-collisions, and incorrect joint dynamics can lead to unrealistic cascading failures. The lack of large-scale annotated datasets has therefore prevented the development of robust learning-based prediction methods for articulated object properties.
 
In summary, this subsection identifies the first key gap: while geometric and kinematic representations of articulated objects have scaled to tens of thousands of models, the physical instantiation of these assets remains largely manual, creating a critical bottleneck for simulation-based robot learning.

\section{Automated Articulated Object Modeling: Structure Without Physics}
 
The complexity and cost of manual curation has motivated recent work on automating articulated object construction. However, existing automated approaches focus on a narrower, more tractable problem: recovering geometric and kinematic structure from incomplete or raw 3D models. These methods operate upstream of the physical property problem.
 
Articulate-Anymesh \citep{qiu2025articulate} is a recent example: given a 3D mesh, potentially with textures, the method segments moving parts by analyzing how geometry changes across a provided video of interaction, estimates which links are articulated and what joint types connect them, and generates realistic textures. The resulting output is a semantic decomposition of the mesh into parts and a kinematic skeleton. This is genuinely useful; it automates a tedious manual step. But it produces a PartNet-Mobility-like \citep{Xiang_2020_SAPIEN} representation: geometry plus kinematics, with physics being left behind.
 
Similarly, Articulate-Anything \citep{le2024articulate} and URDF-Anything \citep{li2025urdf} take multimodal observations such as images, depth maps, object metadata, and semantic annotations from vision-language models, and infer a complete kinematic model: link meshes, joint definitions, joint limits, and optionally initial joint states. These methods are well-suited for real-to-sim scenarios where starting from raw observations is necessary. Yet they too operate at the kinematic level; physical properties are either absent or set to default values. The resulting kinematic models provide a useful starting point, but still require either randomization during training or extensive manual refinement to assign accurate physical parameters before they can be used in realistic manipulation tasks.
 
Infinigen-Articulated \citep{joshi2025proceduralgenerationarticulatedsimulationready} takes a different approach: instead of inferring structure from observations, it procedurally generates articulated assets using category-specific rules and predefined parameter distributions. For example, it might generate a drawer cabinet by composing parameterized box geometries, specifying drawer joints with sampled damping values from a distribution, and rendering the result. This enables large-scale asset generation, and the distributions are tuned to be plausible. However, this approach has significant limitations: (1) extending to new object categories requires manually designing new procedural generators, (2) the generated assets follow a narrow distribution and may not match specific real-world objects, and (3) there is no mechanism for task-specific tuning, as the generated properties are fixed by the predefined distributions rather than adaptive to the user's intended use.
 
Across all these methods, the pattern is consistent: they effectively recover articulated structure but do not address the assignment of physically consistent parameters. This refinement step is where many projects stall: it typically requires both mechanical intuition and simulation expertise, and there is no principled way to validate that the chosen parameters are correct without running full manipulation tasks.
 
This subsection identifies the second key gap: while automated methods have scaled geometric and kinematic asset construction, they have not solved the downstream problem of physically plausible parameter instantiation, leaving a manual gap that their outputs do not bridge.
 
\section{Physical Property Estimation: Representations and Evaluation Gaps}
 
Recent work has explored learning-based approaches for predicting physical properties of 3D objects. These methods aim to directly infer mass, friction, material composition, or other properties from observations, reducing or eliminating manual annotation.
 
DreamPhysics \citep{huang2025dreamphysics} operates on 3D Gaussian Splat (3DGS) representations of objects and scenes and predicts physical properties within a Material Point Method (MPM) simulation framework. The method learns to map visual features from rendered 3D Gaussians to material fields that can be simulated. PhysWorld \citep{yang2025physworld} trains graph neural network-based world models on scenes represented as point clouds, learning to predict deformable object dynamics under MPM simulation. Pixie \citep{le2025pixie} similarly learns to predict material fields and deformation using datasets of Objaverse assets \citep{objaverse} with associated material annotations. These works demonstrate that physical properties, at least in the form of deformability and material, can be inferred from geometric and visual data.
 
However, these approaches face a critical representational mismatch. They operate on scene-level representations such as 3DGS, point clouds, and voxel grids, and produce continuous property fields such as material densities and Young's moduli across the volume. In contrast, robotics simulators like MuJoCo \citep{todorov2012mujoco} and Isaac Sim \citep{NVIDIA_Isaac_Sim} use discrete, object-level and joint-level parameters: uniform mass per link, global friction and damping coefficients per joint, and stiffness values. Bridging this gap from continuous volumetric properties to discrete simulator parameters remains an open challenge. Moreover, these methods focus on predicting material properties (what an object is made of) rather than the interaction properties (friction, damping) that determine simulation behavior.
 
More recently, vision-language models (VLMs) have shown promise for reasoning about physical properties in robotics contexts. PhysObjects \citep{gao2024physically} demonstrates that training a VLM on household objects with associated physical concepts improves long-horizon manipulation planning. Phys2Real \citep{wang2025phys2real} uses VLMs to construct initial estimates of rigid-body properties such as center-of-mass and friction and then refines these estimates via adaptation during manipulation tasks. This approach has succeeded for rigid objects and relatively simple properties (center-of-mass location), but has not addressed articulated systems.
 
The limitations of prior approaches with respect to articulated objects are substantial. First, articulated systems are fundamentally more constrained than rigid objects: incorrect mass distribution or joint parameters can cause self-collisions, kinematic singularities, or unstable cascading failures that break the entire simulation. Second, prior work has treated physical property estimation as a prediction problem: given observations, output property values. However, there is no mechanism to validate whether the predicted properties produce physically consistent behavior when deployed in a simulator. Vision-language models may generate plausible property values that, when combined, result in unstable or unrealistic dynamics. As a result, existing methods operate as black-box predictors without closed-loop feedback or validation.
 
This subsection identifies the third key gap: while learning-based physical property estimation has made progress on rigid objects, methods either operate on incompatible representations such as continuous fields versus discrete parameters, focus on narrow property classes such as material versus dynamics, or lack mechanisms to ensure that predicted parameters produce physically valid simulation behavior.
 
\section{Synthesis: The Interaction-Readiness Problem}
 
The prior work reviewed above reveals a consistent pattern: the pipeline for creating simulation-ready articulated objects remains fragmented and manual. Geometric and kinematic assets scale easily; tens of thousands of models are available. Physical property estimation methods exist but operate on incompatible representations or assumptions, such as rigid objects, continuous fields, or no validation. Automated articulated object construction methods reduce one form of manual effort (link segmentation and joint prediction) but defer the harder problem (physical parameter instantiation).
 
The result is an interaction-readiness gap: objects may be geometrically complete and kinematically correct, but they cannot be reliably used for robot manipulation without substantial additional work. This gap manifests as a blocking problem in practice: researchers and practitioners often default to simpler rigid-object manipulation tasks (which require only mass and friction) rather than articulated-object tasks (which require careful tuning of dozens of coupled physical parameters). Large-scale manipulation benchmarks reveal this bottleneck in practice. RoboLab \citep{yang2026robolab} avoids articulated object manipulation altogether, sidestepping the problem. RoboCasa365 \citep{robocasa365} does include articulated objects, but only through significant manual effort to validate and tune physical parameters for each asset.
 
This thesis addresses this gap through two complementary contributions. First, we propose a quantitative evaluation framework for interaction-readiness, a formal notion of whether an object asset is suitable for manipulation. This framework decomposes interaction-readiness into measurable components such as stability, semantic alignment, behavioral fidelity, and realism, enabling systematic comparison of object quality independent of any single downstream task. Second, we present a method for automatically generating interaction-ready articulated objects by combining the physical reasoning capabilities of vision-language models with iterative feedback from a physics simulator. This approach integrates the geometric and kinematic assets produced by prior methods with realistic physical parameters, bridging the gap between asset construction and deployment.

\chapter{AN INTERACTION-READINESS EVALUATION PROTOCOL\\FOR PHYSICS-BASED SIMULATION} 
\label{ch:evaluation-protocol}
\section{Introduction}
\label{sec:3.1intro}

The challenge of assessing interaction-readiness is that it is a multifaceted property: an articulated asset can be geometrically valid, kinematically correct, and physically plausible in isolation, yet still fail in practice when deployed for robotic manipulation. Traditional evaluation approaches have relied on downstream task success as a proxy for quality. However, this approach conflates the quality of the asset with the quality of the learning algorithm, the success definition, and the task design. A poorly designed asset might occasionally enable task success through policy adaptation, or conversely, a high-quality asset might fail with a suboptimal learning setup. A more principled approach is to decompose interaction-readiness into measurable components, each capturing a different aspect of what makes an asset suitable for manipulation:
 
\begin{itemize}
\item \textbf{Physical robustness}: Does the asset remain stable when instantiated in simulation without external intervention, and do its default configurations avoid self-collisions and spurious drift?

\item \textbf{Semantic accuracy}: Does the asset match expert expectations about size and initial joint configuration, and does it satisfy user specifications about desired object state?

\item \textbf{Behavioral fidelity}: Do the asset's dynamics match real-world object behavior well enough that policies trained on real data can succeed in manipulation tasks in simulation?

\item \textbf{Visual realism}: Do the asset's dynamics appear physically plausible to external observers?

\item \textbf{Interaction realism}: Do human evaluators perceive the asset's behavior as consistent with real objects when interacting with it in simulation?

\item \textbf{Learning feasibility}: Can downstream policies learn to manipulate the asset effectively, reliably articulating target joints and achieving task objectives?
\end{itemize}

These dimensions are chosen to capture distinct and complementary failure modes observed in simulation-based manipulation, including numerical instability, semantic misalignment, and unrealistic interaction behavior. This motivates an evaluation protocol that operationalizes interaction-readiness as a structured set of measurements, enabling systematic comparison of asset generation and refinement methods.

In practice, the quality of an interaction-ready asset is the end objective of any method, but comparing methods remains difficult without a consistent evaluation framework. Physical stability, semantic correctness, and interaction behavior are often assessed qualitatively or through task-specific metrics, leading to inconsistent conclusions. The proposed protocol addresses this by standardizing evaluation along a common set of dimensions, enabling direct and controlled comparison across methods.

At the same time, the protocol remains flexible. Different applications may emphasize different aspects. For example, sim-to-real transfer may depend more heavily on behavioral fidelity, while task-driven settings may prioritize learning feasibility. Rather than prescribing a single scalar metric, we provide a structured set of measurements that can be selectively applied depending on the downstream use case.

\section{Physical Robustness: Quantifying Physical Equilibrium and Simulation Stability}
\label{sec:stability}
A necessary condition for interaction-readiness is that the asset must be physically stable when initialized in a simulator. Instability at rest is a surprisingly common failure mode: articulated objects with poor mass distributions, misaligned joint axes, or incorrect default configurations may exhibit self-interpenetration, spontaneous drift, or oscillatory behavior even without external forces. These issues often render the asset unusable before any interaction begins.
 
Physical robustness can be assessed through three complementary metrics: penetration depth, position/orientation stability, and joint oscillation. Together, these metrics characterize whether the asset's default configuration is physically plausible and whether the simulation will remain stable during the settling phase before interaction.
 
\textbf{Penetration Depth.}
A fundamental requirement is that the asset's link geometries do not interpenetrate when instantiated in the default configuration. One can quantify mesh interpenetration by aggregating penetration depths reported by the physics engine at contact points. For a joint configuration $\mathbf{q}$, the articulated asset of interest can be loaded and stepped through the simulator to obtain the set of contact manifolds $\mathcal{C}(\mathbf{q})$. Each contact point $p \in \mathcal{C}(\mathbf{q})$ provides a signed separation $s(p)$, with $s(p) < 0$ indicating penetration. The penetration score is defined as
\begin{equation}
\label{eq:penetration-score}
\Phi(\mathbf{q}) \;=\; \sum_{p \in \mathcal{C}(\mathbf{q})} \max\!\left(0,\,-s(p)\right),
\end{equation}
which equals the total penetration depth summed over all penetrating contact points. Larger values of $\Phi(\mathbf{q})$ indicate more severe interpenetration, which typically leads to unstable simulation due to large corrective contact impulses. Assets with zero or near-zero penetration are more likely to simulate stably.
 
\textbf{Position/Orientation Stability.}
Beyond initial interpenetration, one should assess whether the asset drifts or tumbles after being instantiated. For this purpose, let $\mathbf{q}_0$ denote the initial joint configuration used to instantiate the articulated asset at reset. Starting from $\mathbf{q}_0$, the simulator can be run for a settling period $T_{\mathrm{set}}$ to allow contacts and passive joints to reach a static equilibrium. The root pose of the object can be recorded as a reference state $(\mathbf{x}_{\mathrm{ref}}, \mathbf{R}_{\mathrm{ref}})$, and the simulation continues for an additional horizon $T_{\mathrm{test}}$ with no external actuation (zero actions). Let $(\mathbf{x}_t, \mathbf{R}_t)$ be the root position and orientation during this evaluation window. The stability deviations are defined as the maximum drift from the reference:
\begin{align}
D_{\mathrm{pos}}(\mathbf{q}_0) 
&= \max_{t \in [0, T_{\mathrm{test}}]} \left\lVert \mathbf{x}_t - \mathbf{x}_{\mathrm{ref}} \right\rVert_2, \\
D_{\mathrm{ori}}(\mathbf{q}_0) 
&= \max_{t \in [0, T_{\mathrm{test}}]} d\!\left(\mathbf{R}_t, \mathbf{R}_{\mathrm{ref}}\right).
\end{align}
An asset is considered to pass the stability test if both deviations remain below predefined thresholds. $D_{\mathrm{pos}}(\mathbf{q}_0) \le 10^{-3}\,\mathrm{m}$ and $D_{\mathrm{ori}}(\mathbf{q}_0) \le 10^{-2}\,\mathrm{rad}$ can be such representative values. These thresholds correspond to sub-millimeter positional drift and small angular deviations, ensuring that the object remains effectively stationary after initialization. Low values indicate that the asset is numerically stable and does not exhibit spurious motion in the absence of external forces, which is critical for reliable tabletop manipulation.
 
\textbf{Joint Oscillation.}
A third stability concern is underdamped joint dynamics, which cause uncontrolled oscillation around equilibrium. During the evaluation window after settling, joint trajectories $q_{j,t}$ for each joint $j$ can be recorded. Oscillatory behavior is characterized using (i) the motion amplitude $A_j = \max_t q_{j,t} - \min_t q_{j,t}$ and (ii) direction reversals in discrete joint velocities $\Delta q_{j,t} = q_{j,t} - q_{j,t-1}$. An asset is marked as exhibiting problematic oscillation if there exists any joint $j$ that exhibits at least three sign changes in $\Delta q_{j,t}$ and has $A_j > \epsilon$ (where $\epsilon$ is a small amplitude threshold, e.g., 0.05 radians for revolute joints). Such oscillation often reflects underdamped or poorly tuned joint dynamics and indicates that the asset's parameters need refinement. Conversely, joints that settle smoothly without oscillation suggest well-tuned damping.
 
Together, these three metrics provide a quantitative characterization of physical robustness and can serve as early-stage quality checks before pursuing interaction-based evaluation.

\section{Semantic Accuracy: Benchmarking Alignment with Human Physical Intuition}
\label{sec:alignment}
Beyond numerical stability, an interaction-ready asset should align with human expectations about the object's size and default configuration. An object that is 10 times too large or begins in a physically implausible state (e.g., a stapler body intersecting its own lid) will fail in practice even if it is numerically stable.
 
One approach to assessing semantic accuracy is to compare against expert-curated reference assets. Benchmarks such as GenSim2 \citep{huagensim2} provide manually specified object scales and default joint configurations, developed by practitioners with domain expertise. While these reference assets may not have fully specified physics parameters, their scale and initial configurations typically reflect reasonable defaults suitable for task execution. Deviations from these references can indicate issues in the generated asset.
 
\textbf{Scale Deviation.}
One can measure the normalized difference between the object scale of a candidate asset and the corresponding reference asset. Let $s_{\text{ref}}(o)$ denote the scale of object $o$ in the reference and $s_{\text{candidate}}(o)$ the scale of a candidate variant. The scale deviation is defined as
\begin{equation}
\Delta_{\text{scale}}(o) = \frac{\big| s_{\text{candidate}}(o) - s_{\text{ref}}(o) \big|}{s_{\text{ref}}(o)}
\label{eq:scale_error}
\end{equation}
Small deviations (e.g., less than 10\%) indicate that the candidate asset has obtained a reasonable size. Large deviations suggest that the asset generation process may have failed to infer the object's true dimensions.
 
\textbf{Default Joint Configuration Deviation.}
Similarly, one can assess whether the asset's initial joint configuration matches that of the reference. For articulated objects with a primary task-relevant joint, the joint configuration deviation can be defined as
\begin{equation}
\Delta_{\text{joint}}(o) = \big| q_{\text{candidate}}(o) - q_{\text{ref}}(o) \big|
\label{eq:joint_error}
\end{equation}
where $q$ denotes the initial position (in radians) of the task-relevant joint. Small deviations indicate faithful reproduction of the reference configuration, while large deviations may indicate misalignment.
 
\textbf{User Specification Alignment.}
In addition to reference-based metrics, one may evaluate whether a generated asset satisfies explicit user requirements specified during the generation process. For instance, if a user requests ``a slightly open lid'' or ``a very small trashcan,'' one can assess whether the resulting asset matches these specifications. This evaluation is necessarily qualitative, conducted through visual inspection of the asset in simulation.
 
Each asset is assigned a prompt alignment score $a(o,v) \in [0,1]$, which reflects the degree to which the generated object satisfies the user-specified properties for object $o$ under variant $v$. A score of $1$ indicates full alignment, where both the object scale and articulated state match the specification. Intermediate values correspond to partial alignment (e.g., correct scale but incorrect joint configuration, or vice versa), while a score of $0$ indicates failure to satisfy the specification (e.g., conflicting joint states or grossly incorrect scale).

The mean prompt alignment for each variant is then computed by averaging scores across all objects:
\begin{equation}
    A(v) = \frac{1}{N}\sum_{o=1}^{N} a(o,v)
\end{equation}
where $N$ is the number of objects. In practice, we can report $A(v)$ as a percentage for ease of comparison across methods.
 
User specification alignment captures semantic faithfulness that reference-based metrics may miss. An asset can be stable and match reference scales yet violate user intent (e.g., correct physics but wrong lid angle), or match reference defaults while contradicting user guidance. Evaluating this dimension ensures that refinement processes preserve semantic fidelity to user specifications, independent of simulator correctness.

\section{Behavioral Fidelity: Probing Predictive Realism via Zero-Shot External Policies}
\label{sec:vla-eval-method}

A fundamental desideratum for interaction-readiness is that an articulated asset's dynamics must encapsulate real-world physical priors with sufficient high fidelity such that control policies conditioned on real-world distributions can generalize to the simulated manifold without explicit retraining. This requirement motivates our formulation of behavioral fidelity as a formal evaluative dimension. We hypothesize that if a zero-shot policy, optimized on a diverse corpus of real-world interactions, successfully manipulates an asset in simulation, it serves as a strong proxy for the convergence of the asset's underlying physical parameters toward a physically realistic equilibrium. 

\subsection{Methodological Framework: The Class of Exogenous Probes}

To operationalize this, we propose the use of \textit{exogenous policy probes}—independent, pretrained models that possess an invariant representational prior of causality and motion derived from vast datasets of real-world robotic manipulation. Within this framework, contemporary Vision-Language-Action (VLA) models represent a prominent instantiation of such exogenous probes, serving as a scalable means to operationalize this evaluative class. These models act as exogenous physical oracles, offering a distinct advantage over standard reinforcement learning agents: they do not over-fit to simulation-specific artifacts or unstable contact dynamics, but rather attempt to interact with the object based on learned terrestrial physics.

\subsection{Empirical Manifold Alignment and Correlation}

The quantification of this fidelity is governed by the alignment between simulated and empirical performance manifolds. We define a success rate $S$ for each asset across a predefined number of independent deployments. Concretely, for each episode, we specify a desired object state $\mathbf{s}^\star$ and declare success if the final state $\mathbf{s}_T$ satisfies $\lVert \mathbf{s}_T - \mathbf{s}^\star \rVert \le \epsilon$, where the norm and threshold $\epsilon$ are environment-dependent. To benchmark these results, the probe is deployed on real-world counterparts to record an empirical baseline. The correspondence between these domains is rigorously evaluated using the Pearson product-moment correlation coefficient, designated as the Sim-to-Real Correlation Coefficient (SRCC) \citep{kadian2020sim2real, aljalbout2025reality}. For a set of $n$ assets, given the simulation success rates $S_{sim,i}$ and real-world success rates $S_{real,i}$, the SRCC is formulated as:

\begin{equation}
SRCC = \frac{\sum_{i=1}^{n} (S_{sim,i} - \bar{S}_{sim})(S_{real,i} - \bar{S}_{real})}{\sqrt{\sum_{i=1}^{n} (S_{sim,i} - \bar{S}_{sim})^2} \sqrt{\sum_{i=1}^{n} (S_{real,i} - \bar{S}_{real})^2}}
\end{equation}

High SRCC values indicate that the simulated dynamics possess high predictive utility, effectively bridging the reality gap. Conversely, systematic failures in the probe's execution—manifesting as unnatural oscillations or unrealistic impulsive forces—expose deficiencies in the asset's physical instantiation. By framing evaluation around exogenous probes, we decouple the intrinsic quality of the object modeling from the idiosyncrasies of task-specific policy optimization, ensuring a principled and generalized measurement of interaction-readiness.

\section{Learning Feasibility: Assessing Interaction-Readiness via Physical Exploration under Policy Optimization}
\label{sec:RL-theory}
An important validation of interaction-readiness is the degree to which an asset facilitates the convergence of downstream control policies towards the correct indended behavior instead of exploiting simulation suboptimality. While a perfectly accurate physical model remains a theoretical ideal, a functional asset must, at minimum, allow a policy to achieve task objectives through autonomous exploration. This metric, which we term \textit{Learning Feasibility}, quantifies whether the asset’s dynamic parameters are sufficiently well-conditioned to avoid inducing deceptive local minima or task-irrelevant noise during the optimization process.

\subsection{Methodological Framework}
To assess feasibility, the candidate asset is instantiated within a physics-based simulation manifold where a downstream agent is trained via Reinforcement Learning (RL). We utilize a state-based policy $\pi_\theta$, optimized using online RL, conditioned on an articulation-centric reward signal. By fixing a sample budget $\mathcal{B}$ (e.g., $3 \times 10^3$ iterations), we use the final success rate as a proxy for the asset's "learnability." An asset is considered interaction-ready if it enables a stable gradient signal, allowing the policy to traverse the transition from stochastic exploration to purposeful manipulation.

\subsection{Structured Reward Decomposition}
To ensure the evaluation is grounded in fundamental physical interaction rather than idiosyncratic reward exploitation, we propose using a formal staged articulation-centric reward function $\tilde{r}_t$. Below is an example articulation-centric reward function to train a Franka Emika (Panda) arm equipped with a Robotiq 2F-85 adaptive gripper. The following formulation is intended to be generalizable to any robotic system and articulated asset distribution with minimal adaptation.

\paragraph{Geometric Guidance and Planar Alignment}
Let $p_t^{\mathrm{ee}} \in \mathbb{R}^3$ and $p_t^{\mathrm{obj}} \in \mathbb{R}^3$ represent the end-effector (EE) and object coordinates, respectively. We define the Euclidean distance $d_t = \|p_t^{\mathrm{ee}}-p_t^{\mathrm{obj}}\|_2$ and planar distance $d_t^{xy} = \|(p_t^{\mathrm{ee}}-p_t^{\mathrm{obj}})_{xy}\|_2$. The policy is first incentivized through a smooth reaching objective:
\begin{equation}
r_t^{\mathrm{reach}} = \omega_1 \cdot \left(1 - \tanh\left(\frac{d_t}{\sigma_1}\right)\right)
\end{equation}
Complementing this, a planar alignment term encourages horizontal approaching over the interaction region, which in practice facilitates the discovery of optimal contact manifolds before fine articulation:
\begin{equation}
r_t^{\mathrm{reach\_xy}} = \omega_2 \cdot \left(1 - \tanh\left(\frac{d_t^{xy}}{\sigma_2}\right)\right)
\end{equation}

\paragraph{Proximity-Gated Articulation Progress}
For an object with $d$ joints, we define the normalized joint state $x_t \in [0, 1]^d$, where each coordinate $x_{t,i} = (q_{t,i} - l_i) / (u_i - l_i)$ is scaled by the asset's runtime joint limits. Given a target state $x_{\mathrm{target}}$, the articulation progress is measured by the $L_1$ error, $e_t^{\mathrm{joint}} = \|x_t - x_{\mathrm{target}}\|_1$. Crucially, to isolate purposeful interaction from far-field motion, the reward is \textit{explicitly gated} by a proximity boolean $\mathbf{1}[d_t < \delta]$:
\begin{equation}
r_t^{\mathrm{joint}} = \mathbf{1}[d_t < \delta] \cdot \omega_3 \cdot \exp(-\alpha \cdot e_t^{\mathrm{joint}})
\end{equation}
This gating mechanism ensures that the policy receives articulation-progress signals only when the EE is within the interaction envelope, thereby suppressing spurious rewards arising from numerical drift or unmodeled vibrations in the joint limits during distal movement.

\paragraph{Physical Regularization}
To bias the policy toward physically plausible trajectories and approach vectors, we introduce the following penalty terms:
\begin{align}
r_t^{\mathrm{pen\_grip}} &= -\lambda_1 o_t \\
r_t^{\mathrm{pen\_down}} &= -\lambda_2 (1 - \langle z_t^{\mathrm{ee}}, [0,0,-1] \rangle) \\
r_t^{\mathrm{pen\_xy}}   &= -\lambda_3 d_t^{xy}
\end{align}
where $o_t \in [0,1]$ denotes the gripper-open fraction and the inner product penalizes deviations from a vertical downward approach. The total per-step reward is scaled by the control timestep $\Delta t$ to maintain magnitude consistency across varying control rates:
\begin{equation}
r_t = \Delta t \cdot \left( r_t^{\mathrm{reach}} + r_t^{\mathrm{reach\_xy}} + r_t^{\mathrm{joint}} + \sum r_t^{\mathrm{penalties}} \right)
\end{equation}

\subsection{Checkpoint Selection and Success Quantification}
To mitigate the impact of reward exploitation or stochastic instability, one can utilize a separate success metric $S$ for intra-asset checkpoint selection:
\begin{equation}
S = \text{clip}\left( 1 - \frac{\|x_t - x_{\mathrm{target}}\|_1}{\max(\|x_0 - x_{\mathrm{target}}\|_1, \epsilon)}, 0, 1 \right)
\end{equation}
This metric is primarily an \textit{intra-asset} criterion. Because the normalization is contingent on the asset’s own (potentially corrupted) joint metadata, absolute success values are not directly comparable across the object distribution. In other words, for such assets, the metric reflects how well a policy satisfies the task under that asset's own physical specification, not a universally fair measure of goal attainment across assets. Instead, we select the optimal checkpoint per asset based on peak success metric and perform final assessment via rollout reliability tests to determine the true degree of interaction-readiness.

\section{Visual Realism: Foundation Models as Objective Judges of Motion Fidelity}\label{sec:vlm-judge}
This dimension of interaction-readiness addresses the discrepancy between analytical stability and physical plausibility. While an articulated asset may satisfy first-order numerical constraints and facilitate task completion within a specific optimization loop, it may nonetheless exhibit higher-order kinematic and dynamic inconsistencies. Such artifacts—including latent causal violations, non-Newtonian impulse transfer, or temporal discontinuities in joint dynamics—often satisfy the simulator's numerical solver while remaining perceptually and functionally inconsistent with terrestrial physical priors. We specifically aim to capture these nuanced deviations, such as telekinetic motion prior to contact or rigid bodies exhibiting non-characteristic material deformation. While human observation remains the gold standard for identifying such discrepancies, manual annotation is inherently unscalable for the systematic benchmarking of large-scale, automated asset distributions.

To resolve this scalability bottleneck, we adopt a \textit{VLM-as-a-Judge} protocol, utilizing Vision-Language Models (VLMs) as automated evaluative instruments for physical reasoning as established in \citet{le2024articulate} and \citet{le2025pixie}. Large-scale VLMs, having been pretrained on internet-scale video corpora containing vast distributions of real-world physical interactions, have internalized an implicit manifold of terrestrial physics. We operationalize the VLM as a zero-shot discriminator to identify non-physical artifacts that traditional scalar metrics fail to characterize. By leveraging these models, we provide a scalable approximation of human perceptual judgment of physical plausibility, ensuring that simulated dynamics align with the causal priors required for high-fidelity modeling.

\subsection{Methodological Framework}
The perceptual fidelity of an asset distribution is quantified via a multi-judge validation protocol designed to mitigate foundation model stochasticity. Realism scores are averaged across a set of canonical manipulation tasks $\mathcal{T}$ using multiple independent VLM architectures. For each task-judge-method triplet, we execute $N$ independent inference trials to capture reasoning variance. The final realism metric $\mathcal{R}$ is computed as:
\begin{equation}
\mathcal{R} = \frac{1}{|\mathcal{T}|} \sum_{t \in \mathcal{T}} \bar{x}_t
\end{equation}
where $\bar{x}_t$ is the mean score for task $t$, evaluated on a consistent scale (e.g., 0--5, as used in prior work~\citep{le2024articulate, le2025pixie}). Standard error is propagated across the task distribution to ensure the assessment is robust to individual model bias.

\subsection{VLM-as-a-Judge Protocol}
The protocol utilizes paired interaction trajectories: a simulated sequence $\mathcal{V}_{sim}$ and a corresponding real-world reference $\mathcal{V}_{real}$. The judge is tasked with identifying specific Newtonian deviations and causal inconsistencies, including but not limited to:
\begin{itemize}
    \item \textbf{Impulse Discontinuities}: Non-physical accelerations or "jumping" artifacts upon contact.
    \item \textbf{Causal Violations}: Telekinetic motion where objects articulate prior to end-effector contact.
    \item \textbf{Material Inconsistency}: Rigid bodies exhibiting underdamped or gelatinous deformation.
\end{itemize}

\begin{tcolorbox}[
  colback=gray!5,
  colframe=black,
  title={VLM-as-a-Judge Prompt Schema (Adapted from \citet{le2025pixie})},
  fonttitle=\bfseries,
  breakable
]
\begin{lstlisting}[basicstyle=\small\ttfamily, columns=fullflexible, breaklines=true]
You are shown candidate videos for the following motion prompt: '{PROMPT}'
Evaluate each candidate for physical realism. Identify violations of causal consistency or impulse discontinuities.

Candidate 0: [video frames]
Candidate 1: [video frames]

Provide your JSON answer following the required schema, specifying the more realistic candidate and the reasoning based on Newtonian physics.
\end{lstlisting}
\end{tcolorbox}

\section{Interaction Realism: Validating Dynamics through Human-Object Teleoperation}
\label{sec:human-telelop}

The terminal dimension of our evaluation protocol addresses the subjective but essential quality of interaction authenticity. Human expert observers possess an internal, high-fidelity generative model of Newtonian physics derived from lifelong embodied experience. We therefore formalize \textit{Interaction Realism} via human-in-the-loop teleoperation to validate emergent physical properties---such as force propagation, contact friction, and joint impedance---that remain inaccessible to purely numerical or vision-only assessments.

\subsection{Methodological Framework}

\begin{table}[ht]
\begin{center}
\small
\renewcommand{\arraystretch}{1.25}
\begin{tabular}{|p{3.0cm}|p{12.5cm}|}
\hline
\textbf{Metrics} & \textbf{Questions \& guidance for testing} \\ \hline

\textbf{1. Scale \& initial joint position} &
\parbox[t]{12.4cm}{
\textbf{Before touching the object:}

1. Look at the object relative to the robot.

2. Does the size of the object look reasonable?

3. Is the object resting naturally on the table or surface?

4. Is the object intersecting the table, or tilted unnaturally?

5. Do the movable parts (lid, drawer, hinge, door, handle, etc.) start in a natural position?

6. Are any parts hanging in the air, passing through other parts, or starting in an impossible configuration?

\textbf{Score based on how physically plausible the object appears.}
\vspace{0.3em}
} \\ \hline

\textbf{2. Configuration stability} &
\parbox[t]{12.4cm}{
\textbf{Do not interact with the object. Wait about 5 seconds and observe.}

1. Does the object move by itself without being touched?

2. Do any joints slowly move, drift, or wiggle on their own?

3. Does the object slide, shake, or rotate slightly even when untouched?

\textbf{An object should remain stable and stationary without external forces.}\vspace{0.3em}
} \\ 
\hline

\textbf{3. Joint realism} &
\parbox[t]{12.4cm}{
\textbf{Now teleoperate the robot and interact with the object. Spend about 1--2 minutes interacting with it.}

Try the following actions:

1. Gently touch or tap the object with the robot gripper.

2. Push the object lightly from the side.

3. Try to move or articulate the object's movable part.

4. Repeat the articulation several times.

While interacting, consider the following:

-- Does the object react naturally when touched?

-- Does the object push back or move in a believable way?

-- Does the joint move according to normal physical intuition of the object?

-- Does the joint feel too stiff, too loose, or unstable?

-- Does the object glitch, clip through itself, explode, or push the robot away unrealistically?

-- Can you successfully articulate the joint multiple times without instability?

\textbf{Evaluate realism based on these interactions.}
\vspace{0.5em}
} \\ \hline

\end{tabular}
\caption{Human Evaluation Framework for Interaction Realism}
\label{tab:human-eval-framework}
\end{center}
\end{table}

This evaluative tier utilizes expert teleoperation of a robotic manipulator to probe the asset’s response to impulsive forces and continuous articulation. This procedure provides a holistic validation of the asset's \textit{interaction manifold}: the complex intersection of geometry, mass distribution, and kinematic constraints. We implement a double-blind protocol where operators interact with randomized, anonymous asset variants, assigning a realism score $\mathcal{R}_H \in [0, 10]$ based on a standardized taxonomy of physical plausibility. To ensure inter-rater consistency and scientific rigor, the protocol follows a three-stage diagnostic procedure. This taxonomy, detailed in Table~\ref{tab:human-eval-framework}, decomposes interaction realism into static plausibility, equilibrium stability, and dynamic articulation fidelity.

\subsection{The Role of Human Judgment as Ground Truth}

The protocol serves as the ultimate validation layer of our evaluation hierarchy. Human judgment integrates disparate physical signals---visual motion, haptic feedback, and causal timing---into a unified assessment of realism. Discrepancies between human and automated metrics (e.g., an asset with high Behavioral Fidelity but low Interaction Realism) expose systemic failures in the simulator's contact solver or material parameterization that numerical solvers are blind to. By establishing this human-centric baseline, we ensure that the asset generation pipeline converges not just toward a numerical local minimum, but toward a functionally authentic representation of reality.

\chapter{GROUNDING THE EMERGENT PHYSICAL REASONING CAPABILITIES OF FOUNDATION MODELS VIA SIMULATOR-IN-THE-LOOP}
\label{ch:method}
\section{Introduction}

The preceding Chapter~\ref{ch:evaluation-protocol} introduced interaction-readiness as an evaluable property of articulated assets. This chapter addresses the problem of constructing such assets from incomplete object models. Given an articulated object with missing or unreliable physical parameters, the goal is to produce a model that behaves plausibly and remains stable under manipulation in simulation. The central difficulty is that the missing quantities are not purely geometric. Mass distributions, inertial tensors, frictional effects, joint damping, and initial articulated states are only weakly determined by static mesh geometry, yet small errors in these parameters can lead to severe simulation artifacts such as self-intersection, spontaneous oscillation, or unrealistic contact response. This motivates an approach that incorporates feedback from the simulator to iteratively refine candidate parameters toward physically consistent outcomes.

We formulate this setting as a grounded physical reasoning task. A foundation model provides broad priors over object function, material composition, and everyday mechanical behavior, while a physics simulator supplies concrete feedback about whether the resulting asset is physically admissible. As summarized in Figure~\ref{fig:flowchart}, the proposed method, \methodname{}, separates asset construction into three stages: extracting a structured multi-modal representation from the input URDF, synthesizing physically plausible parameter edits through model-guided reasoning and analytic constraints, and refining the initial articulated state through simulator feedback. This decomposition preserves the generality of foundation-model reasoning while anchoring its outputs to the constraints imposed by rigid-body simulation.

\begin{figure*}[ht]
    \centering
    \includegraphics[width=\textwidth]{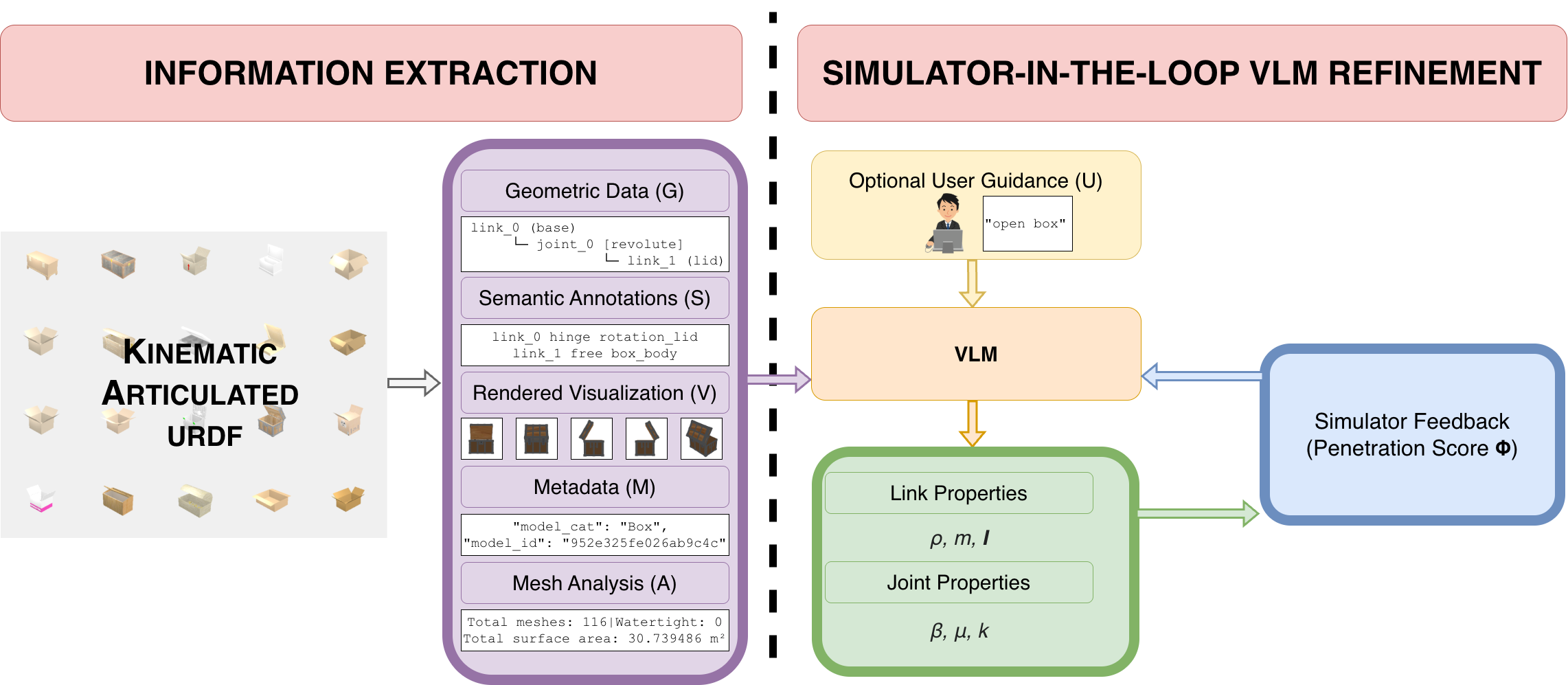}
    \caption{\methodname{} is a simulator-in-the-loop interaction-ready asset creation pipeline with three stages. First, structured information extraction converts an incomplete URDF and its associated meshes into geometric, semantic, visual, and analytic features. Second, VLM-guided property synthesis produces a structured overlay containing scale, link inertial properties, joint dynamics, and optional desired initial state. Third, simulator-in-the-loop state refinement evaluates candidate joint configurations using contact penetration and returns a low-penetration configuration consistent with the requested semantic state.}
    \label{fig:flowchart}
\end{figure*}

\section{Structured Feature Extraction: Preparing High-Fidelity Inputs for Model Reasoning}
The first stage of Figure~\ref{fig:flowchart} converts an incomplete object description in standard URDF format into a multi-modal representation $\mathcal{I}=\{M, G, S, V, \mathcal{A}\}$ that encompasses metadata $M$, geometric data $G$, semantic annotations $S$, rendered visualizations $V$, and mesh analysis $\mathcal{A}$. The purpose of this stage is not merely to serialize the input asset, but to expose the quantities on which physically grounded reasoning must condition: what the object is, how its parts are connected, what those parts appear to be made of, and which geometric measurements constrain feasible physical parameters.

\textbf{Geometric component} $G=\{L,J,\mathcal{M},\mathcal{B}\}$ consists of link properties $L=\{l_1,\ldots,l_n\}$ representing current mass and inertia (if any), kinematic structure $J=\{j_1,\ldots,j_m\}$ representing joint types, limits, and axes, mesh geometries $\mathcal{M}=\{m_1,\ldots,m_n\}$ where each mesh $m_i$ corresponds to link $l_i$, and bounding-box information $\mathcal{B}$. We parse the URDF to extract link structures $L$ and joint definitions $J$. For each link $l_i$, we extract its inertial properties if present, including mass $m_i$ and inertia tensor $\mathbf{I}_i$, along with visual and collision mesh references. Joint definitions include type $\tau_j \in \{\text{revolute}, \text{prismatic}, \text{continuous}, \text{fixed}\}$, parent-child relationships, axis specifications, and limits $[\theta_j^{\min},\theta_j^{\max}]$ for revolute and prismatic joints. The output representation treats joint dynamics as a mandatory triplet $(\beta_j,\mu_j,k_j)$ for every active joint, representing damping, friction, and simulator-supported stiffness; this triplet is generated explicitly even when stiffness is zero for purely passive behavior. This representation preserves the original kinematic constraints rather than re-estimating articulation structure; the subsequent stages refine physical parameters and initial state on top of the existing articulated skeleton.

\textbf{Semantic annotations} $S = \{s_1, \ldots, s_n\}$ are extracted from the URDF's semantics file, where each $s_i$ maps link name $l_i$ to a semantic label (e.g., ``base'', ``lid'', ``hinge'', ``drawer'') and joint type. These semantic mappings provide functional context that informs material inference and dynamics estimation. For instance, a ``hinge'' joint connecting a ``lid'' to a ``base'' suggests different dynamic characteristics than a ``slider'' joint for a ``drawer'' mechanism.

\textbf{Multi-view visual rendering} generates a set of images $V = \{v_1, \ldots, v_k\}$ from $k$ distinct viewpoints in a renderer, 
preserving textures and materials to capture visual cues about material composition. The rendering process accounts for joint configurations and applies proper lighting to highlight material properties. Specifically, we render from canonical viewpoints: front, back, left, right, and perspective projections, ensuring comprehensive visual coverage of the object's geometry and surface properties. The rendered images are provided to the VLM as visual evidence for material inference.

\textbf{Mesh analysis} $\mathcal{A}$ computes volumetric properties for each mesh file in the object's textured mesh directory. For each mesh file, we use Trimesh~\citep{trimesh} to load the geometry and compute: volume $v_i$ (if the mesh is watertight), surface area $a_i$, bounding box dimensions $\mathbf{b}_i = (w_i, h_i, d_i)$, and center of mass $\mathbf{c}_i$ (for watertight meshes). If meshes are not watertight, we use bounding box estimates for each mesh. This distinction is important because many object repositories contain visual meshes that are suitable for rendering but not for exact physical integration. The extracted mesh statistics therefore act as constraints and scale references, not as perfect measurements of material volume.

\section{Physical Property Synthesis: Integrating VLM Priors with Analytical Dynamics}

The second stage of Figure~\ref{fig:flowchart} uses the extracted representation $\mathcal{I}=\{M, G, S, V, \mathcal{A}\}$ to guide the completion of missing physical properties, with the goal of producing an articulated asset whose parameters are both semantically plausible and numerically suitable for simulation.

Specifically, the VLM accepts a prompt containing $\mathcal{I}$ and optional user guidance $U$. Rather than directly rewriting the URDF as unstructured text, the model generates a structured ``overlay'' $\mathcal{O} = \{\mathcal{O}_g, \mathcal{O}_l, \mathcal{O}_j, \mathcal{O}_s\}$ specifying a constrained set of modifications to the original asset. The overlay contains: (1) global modifications $\mathcal{O}_g$ (uniform scale factor $s$ and optional contact/material metadata for downstream use), (2) link modifications $\mathcal{O}_l = \{(l_i, m_i, \mathbf{I}_i)\}_{i=1}^n$ (mass $m_i$, inertia tensor $\mathbf{I}_i$, and optional center of mass for each link), (3) joint modifications $\mathcal{O}_j = \{(j_i, \beta_i, \mu_i, k_i)\}_{i=1}^m$ (damping $\beta_i$, friction $\mu_i$, and simulator-supported stiffness $k_i$ for each joint), and (4) optional initial joint positions $\mathcal{O}_s$ for object state specification. This overlay representation is central to the method: it restricts model output to simulator-relevant parameters, enables deterministic validation before writing the URDF, and preserves the original asset structure whenever no physically justified modification is proposed. We use a single, unified prompt template for all objects. The only variations come from the assets themselves, along with any optional user-provided guidance. See appendix \ref{app:prompt_template} for the prompt template.

The role of the VLM is therefore best understood as providing latent physical priors rather than unconstrained numeric prediction. Visual observations provide evidence about material class and shell thickness; semantic labels provide evidence about functional role; kinematic structure provides evidence about which joints are passive, load-bearing, or return-biased. These sources are fused into parameter proposals, but the final quantities must still satisfy rigid-body constraints. We therefore treat property synthesis as a hybrid inference problem: the VLM estimates the ambiguous latent variables, while analytic formulas and validity checks enforce the mechanical structure required by the simulator.

For each link $l_i$, the geometric analysis provides either a mesh volume $v_i$ when the mesh is watertight, or an approximate volume derived from its bounding box dimensions $\mathbf{b}_i=(w_i,h_i,d_i)$. The VLM is used to infer a material class and an effective density $\rho_i$ from visual appearance, semantics, and optional user guidance. Because many articulated-object meshes represent shells rather than solid bodies, we introduce a hollow factor $\eta_i \in (0,1]$ that captures the fraction of the bounding volume that contributes to mass. The link mass is then estimated analytically as
\begin{equation}
m_i = v_i \rho_i s^3 \eta_i,
\label{eq:vlm_mass_estimate}
\end{equation}
where $s$ is the global uniform scale factor. This separates semantic inference from rigid-body arithmetic: the model supplies the material and structural priors, while mass is constrained by geometry and scale.

Given the estimated mass and scaled dimensions, we construct an inertia tensor $\mathbf{I}_i \in \mathbb{R}^{3 \times 3}$ using simple rigid-body approximations such as boxes or cylinders, with adjustments for observed geometry such as thin plates, elongated handles, or hollow covers. In the ideal rigid-body model, the tensor should satisfy the basic feasibility conditions:
\begin{equation}
\mathbf{I}_i = \mathbf{I}_i^\top, \qquad \mathbf{I}_i \succ 0,
\label{eq:inertia_spd}
\end{equation}
along with consistency of the principal moments. In particular, if $\lambda_{i,1},\lambda_{i,2},\lambda_{i,3}$ are the principal moments of inertia, a physically realizable rigid body must satisfy triangle inequalities such as
\begin{equation}
\lambda_{i,1} + \lambda_{i,2} \geq \lambda_{i,3},
\quad
\lambda_{i,1} + \lambda_{i,3} \geq \lambda_{i,2},
\quad
\lambda_{i,2} + \lambda_{i,3} \geq \lambda_{i,1}.
\label{eq:inertia_triangle}
\end{equation}
These equations define the physical consistency target. In practice, the overlay validation enforces conservative simulator-facing checks: diagonal moments must be positive, and off-diagonal products are removed when they violate basic Cauchy--Schwarz-style consistency bounds. Since the prompt favors simple shape approximations with zero off-diagonal coupling unless strongly justified, these checks keep the final inertial parameters numerically compatible with the simulator even when the original mesh is incomplete or non-watertight.

Joint properties are inferred from joint semantics and visual analysis. For each joint $j_i$, the VLM estimates the full dynamics triplet $(\beta_i,\mu_i,k_i)$ based on joint type $\tau_i$, parent-child semantics, and the apparent function of the mechanism. These parameters are interpreted as passive dynamics by default: damping $\beta_i$ dissipates motion, friction $\mu_i$ resists drift or sliding, and stiffness $k_i$ is set to zero unless spring-like return behavior is intended or strongly supported. A drawer slider, a hinged lid, and a spring-like stapler joint therefore impose qualitatively different constraints on plausible joint dynamics. We require all three quantities to remain nonnegative and finite, and to be consistent with the intended passive behavior of the mechanism. The model also computes the uniform scale factor $s$ to ensure that the asset has realistic dimensions for manipulation tasks while remaining consistent with the extracted geometry.

After generation, the overlay is validated before being applied to the URDF. Links missing from the overlay are assigned conservative minimal inertial values, prismatic joint limits and initial prismatic positions are scaled consistently with the global scale factor, and joint limits are adjusted when necessary so that the documented initial state or simulator default lies inside the feasible range. Because URDF has no universal standard field for default joint position, the desired initial state is retained in the overlay and documented in the revised URDF for downstream simulation setup. This validation step is deliberately mechanical rather than semantic: it does not decide what the object should be, but enforces that the proposed edits define a well-formed articulated rigid-body model.

\section{Closed-Loop State Refinement: Resolving Instabilities via Simulator Feedback}
The third stage of Figure~\ref{fig:flowchart} supports natural-language specification of an object's initial articulated state when it is instantiated within the simulator. Given a user request $U$ (e.g., ``open stapler''), a one-shot VLM can propose a target joint configuration $\mathbf{q}_{\mathrm{VLM}}$. In practice, mesh or kinematic imperfections can make this state self-interpenetrating, which destabilizes physics simulation. We therefore introduce a simulation-in-the-loop state refinement stage that adjusts only the joint configuration to reduce penetration while preserving the intended semantic state.

Let the articulated object have $d$ active joints with configuration $\mathbf{q}\in\mathbb{R}^d$ and joint limits $\{[\theta_k^{\min},\theta_k^{\max}]\}_{k=1}^d$. We define the feasible joint space $\mathcal{Q}=\prod_{k=1}^{d}[\theta_k^{\min},\theta_k^{\max}]$ and initialize $\mathbf{q}^{(0)}=\Pi_{\mathcal{Q}}(\mathbf{q}_{\mathrm{VLM}})$. At each iteration, we simulate the articulated object and compute the penetration score $\Phi(\mathbf{q})$ defined in Equation~\ref{eq:penetration-score}. We treat a configuration as collision-free if the total penetration is below a small tolerance.

For colliding configurations, we localize corrective updates as follows: we group penetrations by colliding link pairs, and then map the most severe collisions to a small set of focus joints $\mathcal{J}_{\text{focus}}^{(t)}$ using the URDF kinematic tree and available part semantics. This localization makes updates efficient and helps preserve the user-specified semantic configuration. Intuitively, we want to reduce penetration without drifting far from $\mathbf{q}_{\mathrm{VLM}}$; otherwise, minimizing penetration alone can converge to a qualitatively different state that does not meet the user's specification.
Conceptually, this refinement seeks a configuration that balances these two objectives,
\begin{equation}
\mathbf{q}^{\mathrm{rev}} \approx \arg\min_{\mathbf{q} \in \mathcal{Q}} \ \Phi(\mathbf{q}) + \lambda \lVert \mathbf{q} - \mathbf{q}_{\mathrm{VLM}} \rVert_2^2,
\label{eq:state_refinement_objective}
\end{equation}
though we do not solve this objective explicitly.

We run an iterative refinement loop as follows. At iteration $t$, we render compact multi-view observations of the current state and query the VLM for an update restricted to the focus joints, using the user request $U$, the static object representation $\mathcal{I}$ (geometry, kinematics, and semantics), the rendered views, and the current penetration score as context. The VLM proposes a candidate update $\Delta \mathbf{q}^{(t)}$ over the focus set, producing
\begin{equation}
\tilde{\mathbf{q}}^{(t+1)} =
\Pi_{\mathcal{Q}}\!\left(\mathbf{q}^{(t)} + \Delta \mathbf{q}^{(t)}\right).
\end{equation}
We accept this update only if it improves physical feasibility:
\begin{equation}
\Phi(\tilde{\mathbf{q}}^{(t+1)}) < \Phi(\mathbf{q}^{(t)}).
\label{eq:acceptance_rule}
\end{equation}
Otherwise, the current state is retained and alternative local proposals are considered. The loop terminates once the penetration score falls below $\varepsilon$ or when a maximum iteration budget is reached, and we return the lowest-penetration configuration encountered.

\section{Robustness of the Refinement Procedure}

The preceding refinement loop can be interpreted as a derivative-free search in a constrained, semantically structured joint space. Unlike standard numerical optimization, the proposal distribution is not purely local in coordinate space: it is informed by rendered observations, collision pairs, joint semantics, and the user's requested object state. This gives the procedure access to high-level physical hypotheses, such as ``a slightly open microwave'' or ``a stapler with spring-like joint behavior'', that are difficult to express using contact gradients alone.

At the same time, VLM proposals are not guaranteed to be monotonically improving. The method therefore separates proposal generation from proposal acceptance. The model may suggest semantically meaningful changes, but the simulator determines whether those changes reduce penetration. This division of labor is essential: foundation-model reasoning supplies candidate directions in an underconstrained physical space, while the simulator supplies a hard feasibility signal. The accepted trajectory $\{\mathbf{q}^{(t)}\}$ is therefore constrained to make non-increasing progress in $\Phi$, except when exploratory proposals are evaluated and rejected.

When refinement stagnates, the search space is progressively relaxed around the same objective rather than reformulated. Nearby perturbations of the current best configuration are explored first, and if these fail, a more systematic local search over the focus joints is used. These fallbacks preserve the semantics of Equation~\ref{eq:state_refinement_objective}: they do not introduce a new notion of object quality, but provide additional coverage of the feasible joint space when the model's first proposals fail to discover a lower-penetration state. In this sense, the simulator-in-the-loop stage converts VLM physical intuition into a verifiable constrained search process.

This structure is particularly useful for articulated objects with tightly coupled kinematics. In such assets, several links may rotate about nearly co-axial or closely spaced joints, so a small joint error can create a large self-intersection. A purely predictive model may correctly infer the semantic state---for example, that a lid should be closed or that a stapler should be open---while still selecting a mechanically infeasible configuration. The refinement loop addresses this ambiguity by treating the VLM proposal as an initialization rather than a final answer. Contact geometry determines which parts of the proposal are physically invalid, and corrective updates are restricted to joints that are implicated in the observed collision structure.

The resulting procedure is therefore not simply a post-processing heuristic. It changes the computational character of asset construction from one-shot prediction to simulate-and-verify search. Candidate states must survive joint-limit projection, semantic-deviation control, and penetration-based acceptance. For complex articulated objects, this makes it less likely that the final state lies in a brittle, near-grazing configuration that appears visually plausible but fails under contact simulation.

The output of Chapter~\ref{ch:method} is thus not merely a revised URDF, but a physically grounded asset construction procedure. The structured overlay turns ambiguous visual and semantic evidence into simulator-facing parameters; analytic constraints ensure that those parameters define a valid rigid-body model; and closed-loop state refinement ensures that the intended initial articulation is compatible with the contact geometry of the asset. Together, these components bridge the gap between foundation-model physical reasoning and interaction-ready simulation.

\section{Engineering Considerations for Scalable Refinement}

The formulation above is intentionally independent of any particular simulator or foundation model. It requires only two abstract interfaces: a simulator that can instantiate articulated assets and report contact geometry, and a multi-modal model that can condition on structured asset information and rendered observations. The concrete implementation choices affect scalability, latency, and cost, but not the conceptual structure of the method.

In our implementation, we use SAPIEN as the simulator backend for the state-refinement loop. SAPIEN provides efficient articulated-object loading, contact diagnostics, and lightweight rendering, which are the operations required by Equations~\ref{eq:state_refinement_objective}--\ref{eq:acceptance_rule}. This makes it suitable for repeated refinement over large asset collections. The same algorithm could be instantiated with another simulator, provided it exposes comparable joint-state control and penetration/contact measurements.

For the VLM component, we use Gemini 2.5 Flash in the scalable refinement pipeline. This choice reflects the operating regime of the algorithm: refinement benefits from fast, repeated, visually grounded proposals rather than a single expensive inference call. Gemini 2.5 Flash provides a practical balance between visual reasoning quality, latency, and API cost, including accessible low-cost/free-tier usage during our experiments. In principle, the method can use any model capable of producing structured overlays and joint-state correction proposals.

\chapter{EXPERIMENTS}
\label{ch:experiments}
\section{Experimental Setup}
\label{sec:exp-setup}
Evaluation of interaction-readiness requires systematic comparison across multiple dimensions of object quality. To enable such comparison, all experiments are conducted within a controlled simulation environment using Isaac Lab 2.3.1~\citep{Mittal_Isaac_Lab_-_2025} built on Isaac Sim 5.1.0~\citep{NVIDIA_Isaac_Sim}, deployed on a workstation equipped with two NVIDIA RTX A6000 GPUs.

The evaluation corpus consists of 11 manipulation tasks spanning 9 articulated objects sourced from PartNet-Mobility~\citep{Xiang_2020_SAPIEN}. The full list of objects, prompts, and associated tasks is provided in Appendix~\ref{app:object-refinement-list}, Table~\ref{tab:task_object_specs}. This includes both Behavioral Fidelity tasks used for VLA evaluation and Learning Feasibility tasks used for policy optimization. All subsequent experiments reference this shared set of objects and task definitions. These tasks are designed to capture two primary modes of interaction. The first consists of grasping and lifting tasks, where the objective is to extract the object from a resting surface; these primarily stress global physical properties such as mass distribution and overall stability. The second consists of articulation tasks, where the objective is to manipulate specific joints to a target configuration; these stress joint dynamics, limits, and self-collision behavior. Together, these categories exercise complementary aspects of interaction-readiness.

For physical property inference in our method, we use Gemini 2.5 Flash as the multi-modal reasoning backbone. This model is selected for its balance between visual understanding capability, inference latency, and accessibility, which is important for iterative refinement.

Finally, results are organized according to the evaluation framework introduced in Chapter~\ref{ch:evaluation-protocol}. For each dimension of interaction-readiness, we first present the corresponding measurements and then interpret these results in terms of whether different asset construction approaches produce objects suitable for realistic and effective manipulation.

\section{Baseline Methodologies and Ablations for Multi-Modal Physical Refinement}
\label{sec:baselinesandablations}
\subsection{Baselines}
We compare the quality of our refined objects and their associated physical properties against three alternative methods.

\textbf{Inverse-Variance Weighted VLM estimates.}
We adapt the VLM-based physical parameter estimation component of Phys2Real~\citep{wang2025phys2real}, extending it from center-of-mass prediction to a full set of URDF-level physical parameters, including global scale, total and per-link masses, per-link inertia tensors, joint damping, stiffness, friction, limits, and initial joint positions. However, we do not adapt their non-VLM estimation techniques. In the original Phys2Real~\citep{wang2025phys2real} framework, VLM priors are fused with interaction-based estimates produced by Rapid Motor Adaptation (RMA) using inverse-variance weighting. The RMA estimate is obtained from an ensemble of adaptation models trained on histories and actions generated by a physics-conditioned reinforcement learning policy in simulation. Training such a policy requires prior knowledge of the downstream task and its associated reward function. 

Since our objective is to evaluate general-purpose articulated asset quality independent of any specific task or environment, incorporating interaction-based adaptation would introduce task-specific inductive bias and leak information about the evaluation setting. We therefore omit RMA-style interaction models for VLM-IVW and rely exclusively on VLM-based physical parameter estimation. In this setting, the uncertainty-aware fusion mechanism of Phys2Real~\citep{wang2025phys2real} reduces to aggregation over VLM predictions alone. While this corresponds to mean aggregation in the original implementation, we instead apply inverse-variance weighting across multi-view, multi-query VLM outputs, treating them as an uncertainty-annotated ensemble.

Following~\citet{wang2025phys2real}, each object is rendered from $N = 5$ canonical viewpoints (front, back, left, right, and perspective), and the VLM is queried $M = 5$ times per view, yielding $N \times M = 25$ independent predictions per parameter. This matches the multi-view sampling used in our \methodname{} pipeline, ensuring a controlled comparison. For each parameter $\theta$, with predictions $\{\theta_i, \sigma_i\}_{i=1}^{N \times M}$, we compute
\begin{equation}
\theta_{\text{IVW}} = \frac{\sum_i \theta_i / \sigma_i^2}{\sum_i 1 / \sigma_i^2}
\end{equation}

This preserves the original Phys2Real fusion formulation and query budget, while exploiting model-reported uncertainty to down-weight inconsistent predictions. We report results using this inverse-variance weighted aggregation of estimated parameters, hereafter denoted as \textbf{VLM-IVW}.

\textbf{Direct VLM (One-Shot).}
As a simpler VLM-only baseline, we remove all multi-query aggregation and treat a single forward pass of the VLM as the final estimate. We use the same refinement prompt and the same $N = 5$ canonical viewpoints (front, back, left, right, and perspective) as in the VLM-IVW baseline, but query the VLM only once with all five rendered images jointly. The model returns a single set of physical parameters and associated uncertainties, which we parse and apply directly to the articulated asset without any further fusion or refinement. We refer to this baseline as \textbf{Direct VLM} in our experiments.

\textbf{Human-Annotated (GenSim2 Articulated Assets).}
As a human-curated baseline, we evaluate using articulated object assets released with GenSim2~\citep{huagensim2}, where physical properties are manually specified by the dataset authors prior to environment generation and manipulation data collection. These assets reflect physically plausible, human-designed parameterizations, while also illustrating the practical challenges of accurately specifying high-dimensional, coupled physical parameters for articulated objects. While GenSim2~\citep{huagensim2} specifies object scale and default configurations, the dataset restricts multi-joint objects to a single joint of interest, and key physical parameters, including joint stiffness and damping, are not modeled. We therefore use simulator default values for all unspecified physical parameters without further refinement, referring to this baseline as \textbf{Human (GenSim2)} in our experiments.

\subsection{Ablations}
\label{sec:ablations}
In addition to the baselines above, we present two ablations evaluated on the same set of objects to further analyze key design choices in our method.

\textbf{Coding Agent with Simulator (Claude Code).} In the first ablation, we evaluate whether physical parameter assignment can be achieved using a general-purpose coding agent with access to a robotics physics simulator. Specifically, we use Claude Code with access to asset rendering via SAPIEN~\citep{Xiang_2020_SAPIEN} and prompt Sonnet 4.6 to assign physical properties. This tests whether access to a simulator alone is sufficient for a coding agent to recover stable and physically consistent assets, in comparison to our structured simulator-in-the-loop refinement procedure. We refer to this ablation as \textbf{Claude Code with simulator} in our experiments.

\textbf{\methodname{} without Semantic Annotations.}
In the second ablation, we remove all semantic annotations from the URDFs and run \methodname{} without providing semantic information to the VLM. This evaluates the extent to which semantic context contributes to physical reasoning and tests the robustness of the method in inferring physically consistent parameters and stable configurations from simulator feedback, geometric, and visual inputs alone. We refer to this ablation as \textbf{\methodname{} (Ours), without semantics} in our experiments.

\section{Physical Robustness and Semantic Fidelity: Assessing Stability and Expert Alignment}
\label{sec:fidelity-eval}
We first evaluate whether each refinement method produces assets that are numerically stable at initialization and semantically consistent with the intended object state. This evaluation combines the physical robustness metrics from Section~\ref{sec:stability} with the semantic accuracy metrics from Section~\ref{sec:alignment}. Physical robustness is measured using penetration depth, position/orientation stability, and joint oscillation. Semantic fidelity is measured using deviation from GenSim2 reference scale and initial joint configuration, together with prompt alignment.

Table~\ref{tab:metrics-table-1} reports these results on the 9-object evaluation set. For penetration depth, all methods produce zero penetration on most objects; nonzero penetration occurs only for the toilet and stapler assets. We therefore report the mean and standard deviation over these two objects, where the metric is informative. Position/orientation deviation and joint oscillation are reported as pass rates over all 9 objects.

\begin{table}[ht]
\begin{center}
\small
\renewcommand{\arraystretch}{1.2}
\setlength{\tabcolsep}{3pt}

\begin{tabular}{
|p{2.9cm}
|>{\centering\arraybackslash}p{2.2cm}
|>{\centering\arraybackslash}p{2.1cm}
|>{\centering\arraybackslash}p{1.9cm}
|>{\centering\arraybackslash}p{2.2cm}
|>{\centering\arraybackslash}p{1.8cm}
|>{\centering\arraybackslash}p{2.0cm}|
}
\hline
\textbf{Method} &
\textbf{Penetration Depth (mm)} &
\textbf{Position / Orientation Deviation Pass Rate (\%)} &
\textbf{Joint Oscillation Pass Rate (\%)} &
\textbf{Scale Difference (\%)} &
\textbf{Initial Joint Difference (rad)} &
\textbf{Prompt Alignment (\%)} \\
\hline

\multicolumn{7}{|c|}{\textit{Baselines}} \\ \hline
Human (GenSim2) & 15.95$\pm$22.55 & 88.9 & \textbf{100} & Reference & Reference & Manual \\ \hline
VLM-IVW & 13.62$\pm$19.26 & 77.8 & 77.8 & 38.35$\pm$31.35 & 0.42$\pm$0.41 & 82.22 \\ \hline
Direct VLM & 61.77$\pm$70.76 & 66.7 & 77.8 & 35.02$\pm$37.58 & 0.50$\pm$0.48 & 86.11 \\ \hline
\methodname{} (Ours) & \textbf{12.83$\pm$18.14} & \textbf{100} & \textbf{100} & \textbf{32.73$\pm$16.51} & \textbf{0.40$\pm$0.48} & \textbf{97.22} \\ \hline

\multicolumn{7}{|c|}{\textit{Ablations}} \\ \hline
Claude Code with simulator & 723.4$\pm$1023.05 & 88.89 & 88.89 & 325.76$\pm$239.56 & 0.48$\pm$0.54 & 52.78 \\ \hline
\methodname{} (Ours), without semantics & 16.01$\pm$22.64 & \textbf{100} & \textbf{100} & 44.48$\pm$42.43 & 0.53$\pm$0.48 & 88.89 \\ \hline

\end{tabular}
\caption{Evaluation on simulation stability. Penetration depth is zero for all refinement methods on all objects except toilet and stapler; reported mean and standard deviation for penetration depth are therefore computed over these two objects only. Position/orientation deviation and joint oscillation are reported as success rates (\%), indicating the fraction of objects that pass the corresponding stability tests. Scale difference and initial joint difference are compared with respect to expert-designed assets from GenSim2.}
\label{tab:metrics-table-1}
\end{center}
\end{table}

\methodname{} achieves the strongest overall physical robustness among the evaluated methods. It obtains the lowest penetration depth among the baselines, while also achieving a 100\% pass rate on both position/orientation stability and joint oscillation. This indicates that the refined assets remain stable after initialization and do not exhibit uncontrolled passive joint motion. In contrast, Direct VLM produces substantially higher penetration depth and lower stability pass rates, suggesting that a single VLM prediction is insufficient to reliably produce physically consistent articulated configurations.

VLM-IVW improves over Direct VLM in penetration depth, but still fails on both position/orientation stability and joint oscillation for a subset of objects. This indicates that multi-query aggregation reduces variance in VLM estimates, but does not enforce global consistency among scale, joint configuration, link geometry, and passive dynamics. The Human (GenSim2) baseline achieves perfect joint-oscillation stability, but does not pass all position/orientation stability tests, reflecting that human-authored assets can still be unstable when imported into the evaluation simulator without additional physical refinement.

The semantic metrics show a similar trend. \methodname{} has the lowest deviation from GenSim2 reference scale and initial joint configuration among the automated methods, while also achieving the highest prompt alignment score. This is important because stability alone is not sufficient: an asset may be numerically stable while violating the user-specified object state or size. The high prompt alignment score indicates that simulator feedback does not simply drive the object toward an arbitrary low-penetration configuration, but preserves the intended semantic state during refinement. A detailed per-object breakdown of prompt alignment scores, together with scoring rationale, is provided in Appendix~\ref{app:prompt-alignment}, Table~\ref{tab:prompt_alignment_breakdown}.

The ablations further clarify the role of structured refinement. Claude Code with simulator access performs poorly on penetration depth, scale difference, and prompt alignment, despite having access to rendered assets through SAPIEN. This suggests that simulator access alone is not sufficient; the refinement procedure must expose the relevant physical quantities and constrain the search over simulator-valid edits. Removing semantic annotations from \methodname{} preserves the two stability pass rates but degrades scale, initial joint, and prompt-alignment metrics. Thus, geometry, visual input, and simulator feedback are sufficient to recover many stable configurations, but semantic annotations improve functional interpretation and alignment with the intended object state.

\section{Behavioral Fidelity: Measuring Predictive Correlation via Zero-Shot VLA Interaction}
\label{sec:vla-eval}
To evaluate behavioral fidelity, we follow the framework defined in Section~\ref{sec:vla-eval-method}, employing three independent Vision-Language-Action (VLA) policies: $\pi_{0}$~\citep{black2025pi0}, $\pi_{0.5}$~\citep{pmlr-v305-black25a}, and GR00T-N1.6~\citep{gr00tn1_2025}. Specifically, $\pi_{0}$-FAST-DROID is a base autoregressive model fine-tuned on DROID~\citep{openpi,khazatsky2024droid}, while $\pi_{0.5}$-DROID is a variant trained with additional knowledge insulation during fine-tuning~\citep{openpi}. Similarly, GR00T-N1.6-DROID is fine-tuned on the DROID dataset~\citep{gr00tn1_2025,khazatsky2024droid}, providing a third independently developed policy with a distinct architecture and training pipeline. All policies are evaluated both in simulation and on corresponding real-world tasks using a Franka robotic arm following the DROID setup~\citep{khazatsky2024droid}, without any additional adaptation or fine-tuning. This paired evaluation enables direct comparison of policy behavior across simulated assets and their corresponding real-world executions.

For simulation evaluation, each policy is deployed for $5$ independent trials per task, and success rate is computed as the percentage of trials in which the desired behavior is achieved, as determined by visual inspection of the final object state. The evaluation is conducted over $11$ tasks, each corresponding to a distinct object--interaction pair defined in Appendix~\ref{app:vla-prompts} (Table~\ref{tab:task-prompts}). Reported success rates are averaged across these tasks.

To obtain the empirical baseline required for SRCC computation, the same policies are deployed on corresponding real-world tasks using the DROID setup~\citep{khazatsky2024droid}, and success rates are computed from recorded executions following the same criteria. SRCC is then computed across tasks by measuring the correlation between simulation and real-world success rates. Table~\ref{tab:VLA-table} summarizes the resulting performance across methods and policies. Detailed per-task results for each policy are provided in Appendix~\ref{app:vla-breakdown} (Tables~\ref{tab:vla-pi0}--\ref{tab:vla-gr00t}).

\begin{table}[ht]
\begin{center}
\small
\renewcommand{\arraystretch}{1.2}
\setlength{\tabcolsep}{3pt}

\begin{tabular}{
|p{3.2cm}
|>{\centering\arraybackslash}p{2.0cm}
|>{\centering\arraybackslash}p{1.4cm}
|>{\centering\arraybackslash}p{2.0cm}
|>{\centering\arraybackslash}p{1.4cm}
|>{\centering\arraybackslash}p{2.0cm}
|>{\centering\arraybackslash}p{1.4cm}|
}
\hline
\textbf{Method} &
\multicolumn{2}{c|}{\textbf{$\bm{\pi}_{\textbf{0}}$}} &
\multicolumn{2}{c|}{\textbf{$\bm{\pi}_{\textbf{0.5}}$}} &
\multicolumn{2}{c|}{\textbf{GR00T N1.6}} \\
\hline
&
\textbf{Simulation Success Rate (\%)} &
\textbf{SRCC} &
\textbf{Simulation Success Rate (\%)} &
\textbf{SRCC} &
\textbf{Simulation Success Rate (\%)} &
\textbf{SRCC} \\
\hline

Human (GenSim2) & 7.27 & 0.40 & 9.09 & 0.32 & 1.82 & 0.08 \\ \hline
VLM-IVW & 16.36 & 0.23 & 20.00 & 0.25 & 10.91 & 0.01 \\ \hline
Direct VLM & 0.00 & N/A & 0.00 & N/A & 1.82 & 0.08 \\ \hline
\methodname{} (Ours) & \textbf{32.73} & \textbf{0.52} & \textbf{45.45} & \textbf{0.68} & \textbf{20.00} & \textbf{0.34} \\ \hline

Real-World Objects & 67.27 & 1 & 67.27 & 1 & 32.73 & 1 \\ \hline

\end{tabular}

\caption{Evaluation of asset fidelity using different policies in real and simulated environments. For VLAs, the policy used in the real and simulated environments is identical.}
\label{tab:VLA-table}

\end{center}
\end{table}

Across all three VLA policies, \methodname{} achieves the highest simulation success rates and the strongest correlation with real-world outcomes. This indicates that the refined assets not only support successful task execution, but also preserve the relative difficulty structure of tasks observed in the real world.

In contrast, Direct VLM performs poorly across all policies, with near-zero success rates and undefined or negligible correlation. This suggests that single-pass parameter estimation fails to produce physically consistent dynamics that can support meaningful interaction, leading to a collapse in both execution and predictive alignment. VLM-IVW improves simulation success rates relative to Direct VLM, but exhibits weak or inconsistent correlation with real-world performance, particularly for GR00T N1.6. This indicates that while aggregation reduces variance in parameter estimates, it does not guarantee that the resulting dynamics reflect real-world interaction priors.

The Human (GenSim2) baseline exhibits moderate correlation for $\pi_{0}$ and $\pi_{0.5}$, but significantly lower performance for GR00T N1.6. This highlights that manually specified assets, while plausible, may not fully capture the physical nuances required for consistent interaction across diverse policy architectures. In contrast, \methodname{} maintains strong correlation across all policies, demonstrating that the refinement process produces dynamics that generalize across different policy families.

Importantly, the gap between simulation and real-world success rates remains substantial across all methods, reflecting the inherent difficulty of the sim-to-real gap. However, the higher SRCC achieved by \methodname{} indicates that its simulated assets provide a more faithful ordering of task difficulty, enabling more reliable prediction of real-world performance.

\section{Learning Feasibility: Evaluating Interaction-Readiness through Policy Learning}\label{sec:rl-eval}
To evaluate learning feasibility, we follow the policy-optimization protocol introduced in Section~\ref{sec:RL-theory}, which measures whether an articulated asset supports stable exploration and learnable task dynamics under a fixed reinforcement learning setup. Unlike the VLA evaluation in Section~\ref{sec:vla-eval}, which probes whether real-world policies transfer into simulation, this experiment evaluates whether a policy trained directly in simulation can discover successful interaction behavior when the asset dynamics define the training environment.

For each object and task defined in Table~\ref{tab:task_object_specs}, we instantiate a corresponding Isaac Lab~\citep{Mittal_Isaac_Lab_-_2025} simulation environment and train an online policy using Proximal Policy Optimization (PPO) \citep{schulman2017ppo} with an asymmetric actor-critic architecture. The policy controls a Franka Emika (Panda) arm equipped with a Robotiq 2F-85 gripper to perform the specified articulation task. All tasks use the same articulation-centric reward formulation based on normalized joint-space progress toward a target configuration, as described in Section~\ref{sec:RL-theory}. Training is performed for 3000 iterations with 4096 parallel environments and no task-specific tuning, so differences in performance primarily reflect differences in the physical properties and interaction dynamics of the underlying assets.

During environment construction, we manually set the initial and target joint configurations for each task to ensure that the relevant joint dynamics could be observed consistently across methods. This step removes certain failure modes related to incorrect initialization (e.g., scale or joint pose), particularly for baseline assets, and therefore results in a comparison that is, if anything, favorable to the baselines. As a result, this evaluation isolates whether the articulated dynamics themselves support stable exploration, contact, and task completion. To ensure that reported success rates reflect meaningful interaction rather than trivial configurations, we also manually inspected policy rollouts across tasks. In particular, we verified that task completion requires non-trivial articulation (e.g., objects are not initialized in already-solved states, often due to incorrect joint limit specifications), and that learned behaviors correspond to physically interpretable interactions rather than degenerate solutions.

Table~\ref{tab:rl-tasks} reports final policy success rates across articulation tasks. \methodname{} achieves the highest average success rate, reaching $97.53\%$, compared to $45.68\%$ for VLM-IVW and $11.11\%$ for both Human (GenSim2) and Direct VLM. The results show that learnability is highly sensitive to articulated dynamics. Direct VLM fails on nearly all tasks, despite the controlled environment setup, indicating that one-shot physical parameter prediction does not reliably produce assets that support stable policy optimization. The Human (GenSim2) baseline also performs poorly on most tasks, reflecting that manually curated assets may provide plausible geometry and task structure while still lacking the dynamic consistency required for reliable articulation learning.

\begin{table}[ht]
\begin{center}
\small
\renewcommand{\arraystretch}{1.2}
\setlength{\tabcolsep}{3pt}

\begin{tabular}{
|p{4.5cm}
|>{\centering\arraybackslash}p{3.4cm}
|>{\centering\arraybackslash}p{2.0cm}
|>{\centering\arraybackslash}p{2.2cm}
|>{\centering\arraybackslash}p{3.2cm}|
}
\hline
\textbf{RL Tasks} &
\textbf{Human (GenSim2)} &
\textbf{VLM-IVW} &
\textbf{Direct VLM} &
\textbf{\methodname{} (Ours)} \\
\hline

Bag Handle Moving & 0 & 0 & 0 & \textbf{100} \\ \hline
Box Closing & 0 & 0 & 0 & \textbf{100} \\ \hline
Cabinet Closing & \textbf{100} & \textbf{100} & \textbf{100} & \textbf{100} \\ \hline
Laptop Folding & 0 & \textbf{100} & 0 & \textbf{100} \\ \hline
Stapler Pressing & 0 & 11.11 & 0 & \textbf{77.78} \\ \hline
Microwave Closing & 0 & 0 & 0 & \textbf{100} \\ \hline
Suitcase Closing & 0 & \textbf{100} & 0 & \textbf{100} \\ \hline
Trashcan Closing & 0 & 0 & 0 & \textbf{100} \\ \hline
Toilet Closing & 0 & \textbf{100} & 0 & \textbf{100} \\ \hline

\hline
\textbf{Average Success Rate (\%)} & 11.11 & 45.68 & 11.11 & \textbf{97.53} \\ \hline

\end{tabular}

\caption{RL policy success rates (\%) across articulation tasks for different asset refinement methods. Each environment uses the same PPO training setup and articulation-centric reward function.}
\label{tab:rl-tasks}

\end{center}
\end{table}

VLM-IVW performs better than Direct VLM on several tasks, suggesting that multi-query aggregation can partially improve the physical conditioning of generated assets. However, its performance is inconsistent: it succeeds on some hinge and prismatic tasks, but fails completely on others. This indicates that aggregation alone can produce task-specific improvements, but does not guarantee broadly learnable dynamics across object categories.

The cabinet task provides an important example of how RL success must be interpreted carefully. All methods achieve $100\%$ success on cabinet closing, but this does not imply that all cabinet assets are physically realistic. As shown below in Figure~\ref{fig:evogen_qualitative_cabinet} in Section~\ref{sec:failure}, several baseline cabinet assets exhibit unstable prismatic joint behavior, including rebound or oscillation after contact. In these cases, the learned policies compensate by repeatedly applying force to counteract these inconsistencies. This demonstrates that RL agents can exploit simulator-specific artifacts to achieve task success, even when the underlying dynamics are physically implausible.

This observation highlights a key limitation of using downstream task success as a standalone metric for asset quality. Learning feasibility measures whether an asset is optimizable under a given simulator, but does not guarantee that the resulting dynamics are physically correct or transferable. For this reason, learnability evaluation have to be complemented by behavioral fidelity (Section~\ref{sec:vla-eval}) and realism assessments (Section~\ref{sec:realism-eval}), which probe whether learned or executed behaviors align with real-world physical priors.


\section{Realism: Cross-Evaluating Motion Plausibility with Human and VLM Judges}
\label{sec:realism-eval}

This section follows the realism evaluations defined in Sections~\ref{sec:vlm-judge}--~\ref{sec:human-telelop}, which assesses motion plausibility through both Vision-Language Model-based evaluation and human-in-the-loop interaction.

For VLM-as-a-judge evaluation, we query two independent vision-language models, Gemini 3.0 and Qwen3-VL 32B, with rendered videos of interaction trajectories. Each model evaluates the physical realism of the observed motion on a scale of 0--5, following the protocol described in Section~\ref{sec:vlm-judge}, which focuses on identifying causal inconsistencies and non-physical dynamics. For each task--method--judge combination, we perform 10 independent trials to account for model stochasticity, and report mean scores with standard error aggregated across tasks. A representative VLM-as-a-judge scoring example and the task prompts used for this evaluation are provided in Appendix~\ref{app:vlm-prompt-examples} (Tables~\ref{tab:vlm_judge_example} and~\ref{tab:task-vlm-prompts}). All interaction trajectories used for VLM-based evaluation are generated using the $\pi_{0.5}$-DROID policy in both simulation and real-world deployments.

For human evaluation, expert operators interact with each asset through teleoperation following the protocol defined in Section~\ref{sec:human-telelop}. Evaluation is conducted under a double-blind setup with randomized asset variants. Each operator assigns a realism score $\mathcal{R}_H \in [0,10]$ based on the structured diagnostic procedure defined in Table~\ref{tab:human-eval-framework}, which evaluates static plausibility, passive stability, and dynamic articulation behavior. Each object is evaluated independently by multiple annotators, and scores are aggregated to produce a mean realism score per method.

\begin{table*}[ht]
\begin{center}
\small
\renewcommand{\arraystretch}{1.2}
\setlength{\tabcolsep}{4pt}

\begin{tabular}{
|p{2.9cm}
|>{\centering\arraybackslash}p{3.2cm}
|>{\centering\arraybackslash}p{4.2cm}
|>{\centering\arraybackslash}p{4.2cm}|
}
\hline
\textbf{Method} &
\textbf{Human Realism Score (0--10) ($\uparrow$)} &
\textbf{Gemini 3.0 VLM-judge Score (0--5) ($\uparrow$)} &
\textbf{Qwen3-VL VLM-judge Score (0--5) ($\uparrow$)} \\
\hline

Human (GenSim2) & 2.18$\pm$0.91 & 1.71$\pm$0.19 & 1.39$\pm$0.14 \\ \hline
VLM-IVW & 3.98$\pm$0.71 & 1.65$\pm$0.19 & 2.43$\pm$0.17 \\ \hline
Direct VLM & 2.82$\pm$1.25 & 2.00$\pm$0.25 & 1.97$\pm$0.17 \\ \hline
\methodname{} (Ours) & \textbf{7.04$\pm$0.54} & \textbf{2.50$\pm$0.21} & \textbf{2.96$\pm$0.13} \\ \hline

\end{tabular}

\caption{Motion realism evaluation using human judgment and VLM-as-a-judge protocols. We report mean $\pm$ standard deviation across 9 articulated objects for human evaluation on a 0--10 scale. For VLM-based evaluation, scores are averaged over 11 tasks with 10 trials per task--judge--method combination on a 0--5 scale, with standard error propagated across tasks.}
\label{tab:human-play-eval}

\end{center}
\end{table*}

The results in Table~\ref{tab:human-play-eval} report values for both human judgments and VLM-as-a-judge protocols. A detailed per-task breakdown of VLM-judge realism scores and per-object human evaluation scores is provided in Appendix~\ref{app:vlm-judge-breakdown} (Table~\ref{tab:vlm_judge_breakdown}) and Appendix~\ref{app:human-eval-breakdown} (Table~\ref{tab:human_eval_breakdown}), respectively. Across all metrics, \methodname{} achieves the highest realism scores, indicating that the refined assets not only satisfy physical constraints but also exhibit dynamics that align with human perceptual expectations.

From the human evaluation perspective, \methodname{} significantly outperforms all baselines, achieving a mean realism score of $7.04$, compared to $3.98$ for VLM-IVW and $2.82$ for Direct VLM. This gap highlights that numerical stability alone is insufficient to guarantee perceptual realism. While VLM-IVW and Direct VLM may produce assets that are partially functional, they frequently exhibit artifacts such as unrealistic joint stiffness, non-physical damping, or implausible contact responses during interaction, which are readily identified by human evaluators.

The VLM-as-a-judge results show a consistent but more nuanced trend. While all methods achieve lower absolute scores than human evaluation, \methodname{} consistently ranks highest across both Gemini 3.0 and Qwen3-VL. The reduced separation between methods suggests that VLM-as-a-judge evaluation is less sensitive to fine-grained interaction artifacts than human judgment. Interestingly, Direct VLM achieves relatively competitive scores under VLM evaluation despite poor performance in behavioral fidelity. This indicates that perceptual realism and functional correctness are not perfectly correlated: an asset may exhibit visually plausible motion while failing to support consistent interaction dynamics. Conversely, VLM-IVW achieves higher scores than the human baseline under Qwen3-VL, suggesting that manually authored assets may lack consistent dynamic tuning, even if their geometric and semantic properties are correct.

Overall, the strong performance of \methodname{} across both human and VLM judges indicates that the refinement process successfully aligns simulated dynamics with both functional and perceptual criteria of physical realism.

\section{Scaling and Generalization: Assessing Quality Across Large-Scale Object Distributions}
\label{sec:scaling-eval}
To evaluate whether the proposed refinement procedure generalizes beyond a small curated set of objects, we conduct a large-scale study on 93 articulated assets spanning 10 object categories. Each asset is initialized from PartNet-Mobility and paired with a natural language prompt, following the same pipeline used in earlier sections. This setting introduces significantly greater variability in geometry, articulation structure, and prompt complexity, providing a stress test for both physical consistency and semantic alignment at scale. The full list of objects and prompts used in this large-scale evaluation is provided in Appendix~\ref{app:large-scale-objects} (Table~\ref{tab:large-scale-objects}).

Table~\ref{tab:metrics-table-large} reports stability metrics on the subset of 80 assets for which all methods successfully produced simulation-ready outputs. Consistent with the small-scale results in Section~\ref{sec:fidelity-eval}, \methodname{} achieves the lowest penetration depth and the highest position/orientation stability pass rate, while matching the strongest baseline on joint oscillation stability. These results suggest that simulator-in-the-loop refinement is a key requirement for scaling articulated asset generation, especially for high-degree-of-freedom objects where parameter coupling becomes more pronounced.

\begin{table}[ht]
\begin{center}
\small
\renewcommand{\arraystretch}{1.2}
\setlength{\tabcolsep}{3pt}

\begin{tabular}{
|p{3.2cm}
|>{\centering\arraybackslash}p{3.0cm}
|>{\centering\arraybackslash}p{4.5cm}
|>{\centering\arraybackslash}p{3.0cm}|
}
\hline
\textbf{Method} &
\textbf{Penetration Depth (mm)} &
\textbf{Position / Orientation Deviation Pass Rate (\%)} &
\textbf{Joint Oscillation Pass Rate (\%)} \\
\hline

VLM-IVW & 18.03$\pm$57.19 & 93.75 & \textbf{98.75} \\ \hline
Direct VLM & 19.86$\pm$60.07 & 77.50 & 82.50 \\ \hline
\methodname{} (Ours) & \textbf{11.65$\pm$37.85} & \textbf{97.50} & \textbf{98.75} \\ \hline

\end{tabular}

\caption{Evaluation on simulation stability across 80 assets successfully refined by all methods (subset of 93 initial assets across 10 object categories).}
\label{tab:metrics-table-large}

\end{center}
\end{table}

However, aggregate metrics alone do not capture the full picture of scalability. Figure~\ref{fig:failure-mode} presents a breakdown of outcomes across all 93 assets, distinguishing successful refinement (passing all numerical stability checks) from several classes of failure, including invalid joint configurations, failure to produce valid parameter estimates, and physically inconsistent simulation behavior.

Notably, failures in joint oscillation are consistently accompanied by failures in position and orientation stability, indicating that these metrics are strongly coupled in practice. As a result, violations of passive stability typically manifest as broader simulation instability rather than isolated effects.

\begin{figure*}[ht]
    \centering
    \includegraphics[width=1.0\textwidth]{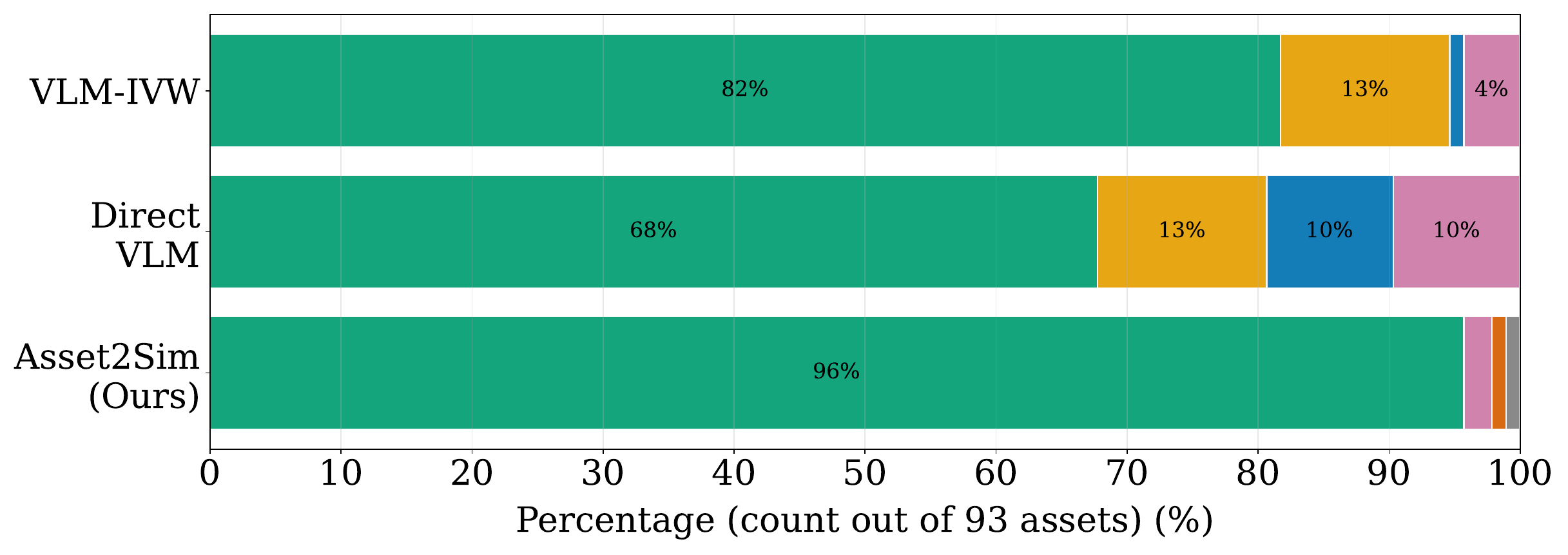}
    \includegraphics[width=\textwidth]{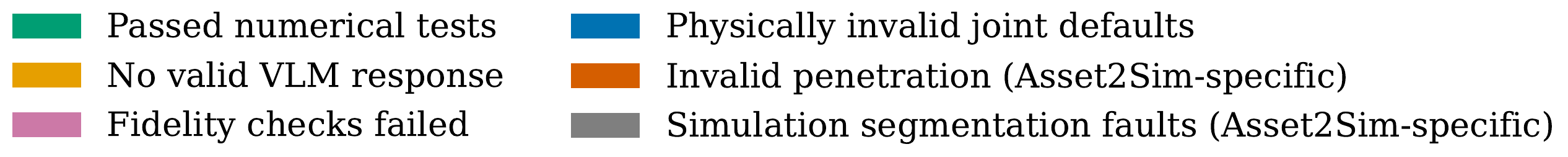}
    \caption{Failure mode distribution across all 93 assets spanning 10 object categories. Each bar reports the percentage of assets that pass numerical stability checks versus those that fail due to different error modes. \methodname{} achieves a 96\% success rate, significantly outperforming VLM-IVW and Direct VLM, which exhibit higher failure rates due to structural inconsistencies and unreliable parameter estimation.}
    \label{fig:failure-mode}
\end{figure*}

The results show that \methodname{} successfully refines 96\% of assets, significantly outperforming both VLM-IVW and Direct VLM. In contrast, Direct VLM exhibits a substantially higher rate of structural failures, including invalid joint configurations and failure to produce simulation-compatible outputs. These failures are particularly pronounced for objects with higher degrees of freedom, such as keyboards, where single-pass estimation struggles to produce coherent parameterizations. More broadly, VLM-only methods lack an explicit mechanism to validate joint defaults and articulation constraints within a simulator, leading to frequent inconsistencies in joint initialization and format. While VLM-IVW improves robustness through aggregation, it remains unable to resolve coupled parameter inconsistencies in a non-trivial fraction of cases.

\methodname{} introduces a small number of failure modes specific to the simulator-in-the-loop refinement process, such as invalid penetration states and segmentation-related issues. These arise from limitations in the underlying simulator representation rather than incorrect physical reasoning. Despite these edge cases, the overall success rate demonstrates that iterative refinement with simulator feedback scales effectively across diverse object distributions.

Finally, while successful refinements (shown in green) satisfy all numerical stability criteria, they do not necessarily correspond to fully interaction-ready assets. Motion realism and task-level usability must be assessed through complementary evaluation protocols, including human interaction, policy-based evaluation, and VLM-as-a-judge assessments, as shown in Sections~\ref{sec:vla-eval}--\ref{sec:realism-eval}.

\section{Failure Analysis: Systematic Limitations in VLM-Guided Physical Refinement}\label{sec:failure}

While the quantitative results in the previous sections demonstrate improvements in stability and interaction performance, they do not fully explain the underlying causes of failure across different methods. To better understand these behaviors, we perform a qualitative analysis of interaction trajectories, focusing on how predicted geometry and physical parameters affect downstream manipulation.

To further characterize these failure modes across different articulation regimes, we analyze three representative object classes: (1) a stapler, which exhibits tightly coupled, near co-axial joints (Figure~\ref{fig:evogen_qualitative_stapler}); (2) a suitcase, which represents a standard revolute joint with contact-driven actuation (Figure~\ref{fig:evogen_qualitative_suitcase}); and (3) a cabinet drawer, which corresponds to a prismatic joint requiring stable linear motion (Figure~\ref{fig:evogen_qualitative_cabinet}). These examples allow us to isolate how errors in joint structure and physical parameterization manifest across qualitatively different forms of articulation.
\begin{figure*}
    \centering
    \includegraphics[width=\textwidth]{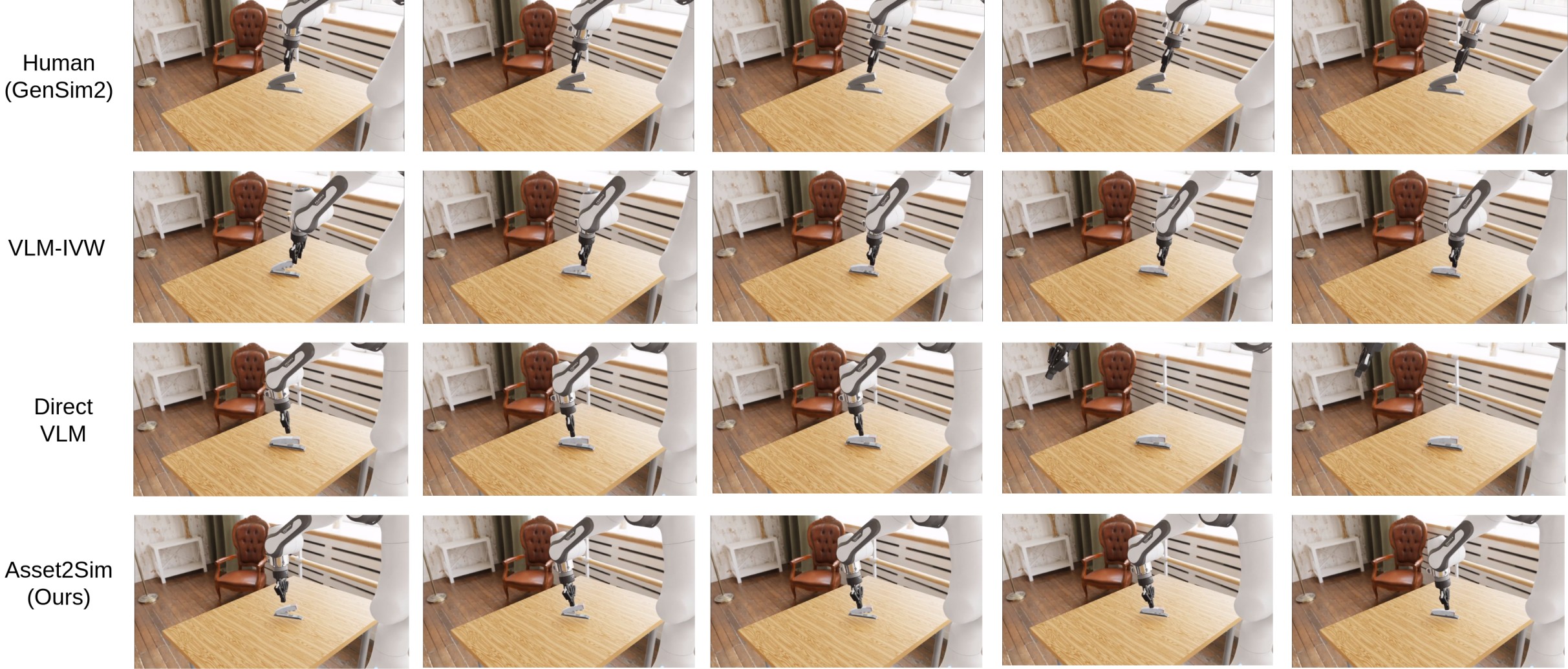}
    \caption{Representative frames (ordered by time) from VLA trajectories in IsaacLab simulations of the stapling task, using URDFs refined by different methods.
The Direct VLM stapler appears in unrealistic configurations; when the gripper makes contact, abnormal internal forces push the gripper away.
The VLM-IVW stapler's joint configuration is unrealistic: the stapler internal body is suspended in midair, while the gripper is able to press it from the side rather than from above.
The VLA tries to press the GenSim2 stapler at multiple contact points and slides upward along the surface, but fails to actuate it.
In contrast, the VLA successfully presses the \methodname{} stapler.}
    \label{fig:evogen_qualitative_stapler}
\end{figure*}

\begin{figure*}
    \centering
    \includegraphics[width=\textwidth]{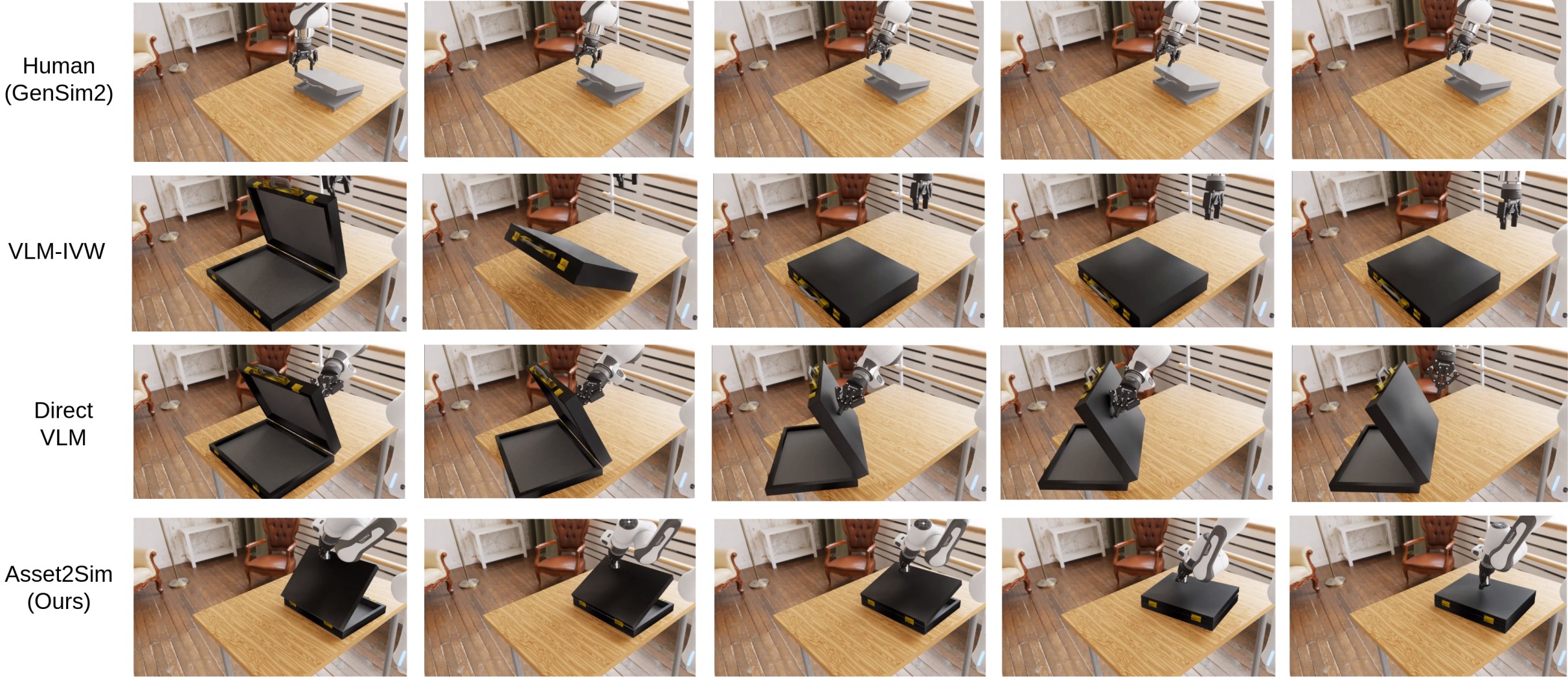}
\caption{Representative frames (ordered by time) from VLA trajectories in IsaacLab simulations of the suitcase closing task, using URDFs refined by different methods. Direct VLM produces an overly stiff articulation: contact fails to actuate the lid, and the entire object translates instead. VLM-IVW yields unstable dynamics, with the lid closing abruptly at initialization due to inconsistent joint parameters. The Human (GenSim2) asset exhibits plausible geometry but remains difficult to actuate, as contact leads to sliding without successful closure. In contrast, \methodname{} produces a physically consistent articulation, enabling successful task completion.}
    \label{fig:evogen_qualitative_suitcase}
\end{figure*}

\begin{figure*}
    \centering
    \includegraphics[width=\textwidth]{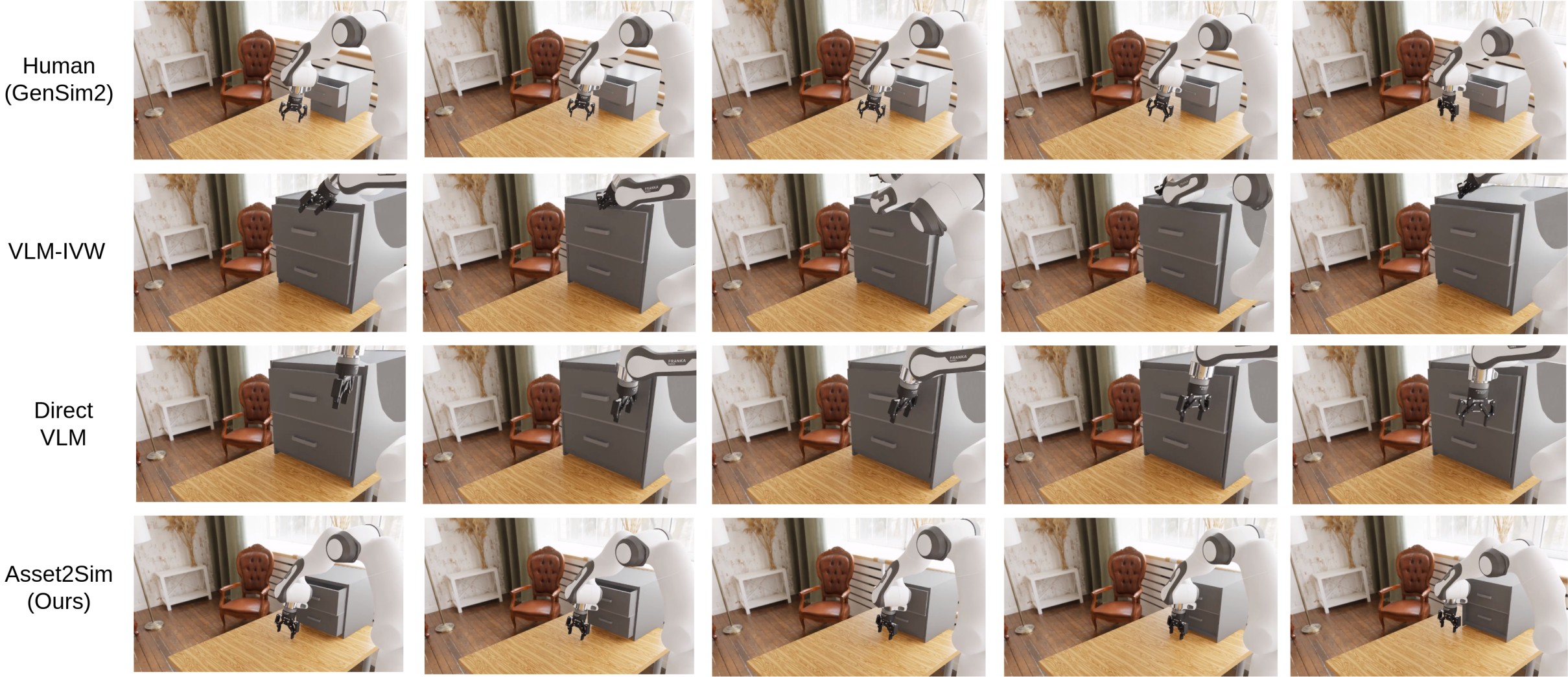}
    \caption{Representative frames (ordered by time) from VLA trajectories in IsaacLab simulations of the cabinet closing task, using URDFs refined by different methods.
All three baselines exhibit unstable prismatic joint behavior. The Human (GenSim2) drawer resists motion and cannot be fully actuated. The VLM-IVW and Direct VLM drawers can be pushed all the way but still exhibits oscillatory behavior, reflecting excessive restoring forces or poorly calibrated damping.
In contrast, \methodname{} drawer is smoothly actuated to the closed position and remains in place after interaction, without observable rebound or oscillation.}
    \label{fig:evogen_qualitative_cabinet}
\end{figure*}

VLM-based baselines exhibit consistent limitations when used to infer articulated object structure and physical properties. First, predicted joint configurations are often geometrically inconsistent, particularly in objects with tightly coupled or nearly co-axial joints (e.g., stapler lid–body, toilet seat–lid). Small errors in joint placement or orientation propagate across connected parts, producing invalid initial states such as misaligned axes, unsupported components, or internal collisions. These inconsistencies lead to unstable objects that cannot support meaningful interaction, as seen in the Direct VLM and VLM-IVW examples.

Second, articulated dynamics are frequently unrealistic due to incorrect or missing physical parameters. In particular, stiffness and damping are often poorly specified, causing joints (e.g., laptop and microwave hinges) to exhibit non-physical behavior such as negligible resistance, excessive compliance, or unstable motion under contact. As a result, even geometrically plausible objects fail to respond correctly during manipulation.

Third, manually constructed baselines mitigate some of these issues by simplifying articulation. In practice, multi-joint objects are often reduced to a single joint of interest to stabilize training and accelerate learning. While this improves tractability, it alters the underlying dynamics and introduces a mismatch between simulated behavior and real-world object structure.

Across these examples, we observe that failure modes are not isolated to specific objects, but arise consistently across different articulation types, including co-axial multi-joint systems, revolute joints, and prismatic mechanisms. This reinforces that the limitations of VLM-based prediction are structural rather than task-specific.

These observations point to a fundamental limitation of prediction-based approaches: VLMs operate in an open-loop setting, where geometry and physical parameters are inferred without validating their consequences under interaction. As a result, small errors in structure or dynamics compound during simulation, leading to failure at the level of contact and actuation.

This suggests that reliable physical reasoning with foundation models cannot be achieved through one-shot prediction alone. Instead, it requires a closed-loop refinement process in which candidate parameters are iteratively evaluated and corrected based on simulator feedback. By explicitly enforcing physical consistency during refinement, such a loop bridges the gap between semantic inference and interaction-ready dynamics.

\section{Efficiency: Autonomous Multi-Modal Refinement vs. Manual Physical Modeling}
\label{sec:efficiency}
Beyond physical fidelity and interaction quality, an important consideration is the efficiency of asset generation. In this context, efficiency is defined along two dimensions: (i) the amount of human effort required to produce simulation-ready assets, and (ii) the computational cost of automated refinement, measured by the number of model queries per asset.

Table~\ref{tab:efficiency-analysis} compares different approaches along these dimensions. Manual curation requires substantial human expertise and iterative trial-and-error, making it neither scalable nor parallelizable. While it can produce high-quality assets, the process is inherently labor-intensive and difficult to reproduce at scale.

\begin{table*}[ht]
\begin{center}
\small
\renewcommand{\arraystretch}{1.2}
\setlength{\tabcolsep}{4pt}

\begin{tabular}{
|p{2.7cm}
|>{\centering\arraybackslash}p{3.5cm}
|>{\centering\arraybackslash}p{2.5cm}
|>{\centering\arraybackslash}p{2.5cm}
|>{\centering\arraybackslash}p{2.5cm}|
}
\hline
\textbf{Method} &
\textbf{Number of Queries per Asset} &
\textbf{No Expertise Needed} &
\textbf{Parallelizable} &
\textbf{Fully Autonomous} \\
\hline

Manual Curation & Human labor & $\times$ & $\times$ & $\times$ \\ \hline
VLM-IVW & 25 & $\checkmark$ & $\checkmark$ & $\checkmark$ \\ \hline
Direct VLM & 1 & $\checkmark$ & $\checkmark$ & $\checkmark$ \\ \hline
\methodname{} (Ours) & 2.26$\pm$3.72 & $\checkmark$ & $\checkmark$ & $\checkmark$ \\ \hline

\end{tabular}

\caption{Efficiency analysis of different methods. The number of queries per asset is reported as the average across 93 assets spanning 10 object categories. For VLM-only baselines, we report their default query counts.}
\label{tab:efficiency-analysis}

\end{center}
\end{table*}

VLM-based approaches eliminate the need for manual intervention and enable parallel asset generation. Direct VLM achieves the lowest query cost by producing a single prediction per asset, but, as shown in previous sections, this comes at the expense of physical consistency and interaction reliability. VLM-IVW improves robustness by aggregating multiple predictions, but requires a significantly higher number of queries (25 per asset), increasing computational cost.

In contrast, \methodname{} achieves a balance between efficiency and reliability. Despite operating in a closed-loop setting with simulator feedback, it requires only $2.26 \pm 3.72$ queries per asset on average, which is an order of magnitude lower than VLM-IVW. This demonstrates that iterative refinement does not necessarily incur high computational overhead, and that a small number of feedback-driven updates is sufficient to correct physically inconsistent predictions.

These results highlight that efficiency and physical validity are not inherently at odds. By integrating multi-modal inference with simulator feedback, \methodname{} maintains low query complexity while producing interaction-ready assets, enabling scalable generation without sacrificing physical realism.

\chapter{CONCLUSION} 
\label{ch:conclusion}

In this thesis, we addressed the problem of constructing interaction-ready articulated objects for physics-based robot simulation. We argued that current pipelines for object modeling remain incomplete: while geometric structure and kinematic relationships can be inferred at scale, the assignment of physically consistent properties required for stable and realistic interaction remains largely unsolved. This gap limits the scalability and reliability of simulation-based robot learning.

To address this, we introduced interaction-readiness as a concrete and measurable objective. We proposed a quantitative evaluation framework that decomposes object quality into physical stability, semantic alignment, behavioral fidelity, learning feasibility, and qualitative realism. This framework enables systematic analysis of object assets independent of downstream task success, and reveals failure modes that are difficult to diagnose through qualitative inspection alone.

Building on this formulation, we presented \methodname{}, a multi-modal, simulator-in-the-loop approach for generating interaction-ready articulated objects from incomplete 3D assets. The method integrates geometric, visual, and semantic information to infer physical properties, and refines these estimates through iterative feedback from a physics simulator. By incorporating simulation into the refinement loop, the approach directly enforces physical consistency, addressing limitations of purely predictive, model-based pipelines.

Through extensive experiments, we demonstrated that improving object quality has a direct impact on downstream outcomes. Objects generated by our method exhibited greater physical stability, more realistic interaction behavior, and improved performance under both policy learning and zero-shot interaction settings. These results suggest that the fidelity of object assets is not merely a secondary concern, but a central factor in enabling scalable and reliable robot learning through physically realistic simulation.

Despite these advances, several challenges remain. First, the proposed evaluation framework, while systematic, introduces additional computational overhead, particularly when incorporating policy learning and model-based evaluation components at scale. Second, the approach depends on the quality of the underlying asset representation; in practice, the method performs best when basic structural and semantic information is correctly specified.

Looking forward, this work opens several directions for future research. One promising direction is the development of more efficient evaluation strategies that preserve diagnostic power while reducing computational cost. Another is improving robustness to incomplete or noisy object representations, potentially through joint inference of structure and physical properties. Finally, extending these ideas to large-scale, automatically generated environments raises questions about how to model object realism jointly with scene composition, enabling interaction diversity and scalable generation of interaction-rich environments.

In summary, this thesis took a step toward closing the gap between large-scale object datasets and physically realistic simulation for robot learning. By formalizing interaction-readiness and introducing a simulator-in-the-loop refinement paradigm, we provided both a framework for evaluating object quality and a practical approach for improving it. We hope this work contributes to enabling more reliable, scalable, and physically grounded simulation for robot learning.

\end{mainf}

\begin{append}

\chapter{\MakeUppercase{Evaluation Objects and Task Definitions}}
\label{app:object-refinement-list}
This appendix specifies the objects, prompts, and tasks used in Chapter~\ref{ch:experiments}. Each object is sourced from PartNet-Mobility \citep{Xiang_2020_SAPIEN} and paired with a natural language prompt for physical property inference.
\begin{table*}[ht]
\begin{center}
\small
\renewcommand{\arraystretch}{1.2}
\setlength{\tabcolsep}{4pt}

\begin{tabular}{
|p{0.3cm}
|p{1.6cm}
|>{\centering\arraybackslash}p{1.6cm}
|p{5.0cm}
|p{3.0cm}
|p{3.0cm}|
}
\hline
\multicolumn{1}{|>{\centering\arraybackslash}p{0.3cm}|}{\textbf{\#}} &
\multicolumn{1}{>{\centering\arraybackslash}p{1.6cm}|}{\textbf{Object}} &
\multicolumn{1}{>{\centering\arraybackslash}p{1.6cm}|}{\textbf{PartNet-Mobility ID}} &
\multicolumn{1}{>{\centering\arraybackslash}p{5.0cm}|}{\textbf{Prompt}} &
\multicolumn{1}{>{\centering\arraybackslash}p{3.0cm}|}{\textbf{Behavioral Fidelity (VLA) Tasks}} &
\multicolumn{1}{>{\centering\arraybackslash}p{3.0cm}|}{\textbf{Learnability (Online RL) Tasks}} \\
\hline

1 & Stapler & 103275 &
I want a slightly open stapler. &
\parbox[t]{3.1cm}{Grasping the stapler \\ Pressing the stapler \vspace{0.3em}} &
Pressing the stapler  \\  
\hline

2 & Box & 100221 &
A very small jewelry box or treasure chest with a slightly open rotational lid. Low damping for smooth hinge behavior. &
\parbox[t]{3.1cm}{Lifting the box \\ Closing the box} &
Closing the box \\ \hline

3 & Microwave & 7236 &
I want a microwave with its door halfway open. Low damping for hinge-like behavior. &
Closing the microwave door &
Closing the microwave door \\ \hline

4 & Toilet & 102703 &
I want a very small toilet for a baby, with its lid fully open. Low damping for hinge-like behavior. &
Closing the toilet lid &
Closing the toilet lid \\ \hline

5 & Laptop & 9912 &
I want a slightly open laptop. Low damping for smooth operation. &
Folding the laptop &
Folding the laptop \\ \hline

6 & Bag & 101673 &
I want a small purse or handbag with the handle moving freely, resting sideways. &
Picking up the bag &
Moving the bag handle sideways \\ \hline

7 & Cabinet & 19179 &
I want a very small two-drawer under-desk office cabinet, with the first drawer closed and the second slightly open, with low damping sliding joints. &
Closing the cabinet drawer &
Closing the cabinet drawer \\ \hline

8 & Suitcase & 103755 &
I want a small business suitcase with its lid slightly open and a low damping hinge. &
Closing the suitcase &
Closing the suitcase \\ \hline

9 & Trashcan & 102181 &
A very small personal trash can with its lid slightly open. The hinge has low damping and holds its position. &
Closing the trashcan lid &
Closing the trashcan lid \\ \hline

\end{tabular}

\caption{Complete specification of objects, prompts, and associated tasks used in all experiments. Behavioral Fidelity (VLA) tasks evaluate zero-shot interaction performance with different VLA models, while Learning Feasibility (RL) tasks assess policy learnability through physical exploration.}
\label{tab:task_object_specs}

\end{center}
\end{table*}

\chapter{\MakeUppercase{Per-Object Prompt Alignment Analysis}}
\label{app:prompt-alignment}

This appendix provides a detailed per-object breakdown of prompt alignment scores for all evaluated methods. For each object, we report the assigned score together with a qualitative explanation of the observed behavior, including scale estimation and articulated joint configuration.

\begin{center}
\small
\renewcommand{\arraystretch}{1.2}
\setlength{\tabcolsep}{4pt}

\begin{longtable}{
|p{5.8cm}
|>{\centering\arraybackslash}p{1.2cm}
|p{8.8cm}|
}

\caption{Per-object prompt alignment scores and scoring rationale. Scores evaluate agreement with the natural language prompt along two dimensions: object scale and articulated joint configuration. Each dimension contributes approximately equally to the final score. Partial credit is assigned when scale is approximately correct or when the articulated configuration is close to the requested state. Aggregate prompt alignment percentages reported in the main text are obtained by averaging these scores across all objects.}
\label{tab:prompt_alignment_breakdown} \\

\hline
\textbf{Method} &
\textbf{Score} &
\textbf{Scoring rationale / observed failure mode} \\
\hline
\endfirsthead

\hline
\textbf{Method} &
\textbf{Score} &
\textbf{Scoring rationale / observed failure mode} \\
\hline
\endhead

\hline
\multicolumn{3}{r}{\textit{Continued on next page}} \\
\endfoot

\hline
\endlastfoot

\multicolumn{3}{|c|}{\textit{Bag}} \\* \hline
\methodname{} (Ours) & 1.00 & Correct scale and handle configuration. \\* \hline
VLM-IVW & 1.00 & Correct scale and handle configuration. \\* \hline
Direct VLM & 1.00 & Correct scale and handle configuration. \\* \hline
Claude Code with simulator & 0.50 & Object is substantially oversized relative to the robot workspace. \\* \hline
\methodname{} (Ours), without semantics & 1.00 & Correct scale and handle configuration. \\ \hline

\multicolumn{3}{|c|}{\textit{Box}} \\* \hline
\methodname{} (Ours) & 1.00 & Correct scale and slightly open lid configuration. \\* \hline
VLM-IVW & 0.75 & Object scale is moderately larger than intended, while lid configuration remains aligned. \\* \hline
Direct VLM & 0.75 & Object scale is moderately larger than intended, while lid configuration remains aligned. \\* \hline
Claude Code with simulator & 0.75 & Object scale is moderately larger than intended, while lid configuration remains aligned. \\* \hline
\methodname{} (Ours), without semantics & 0.75 & Object scale is moderately larger than intended, while lid configuration remains aligned. \\ \hline

\multicolumn{3}{|c|}{\textit{Cabinet}} \\* \hline
\methodname{} (Ours) & 1.00 & Correct scale and drawer configuration. \\* \hline
VLM-IVW & 0.75 & Object scale is moderately larger than intended, while lid configuration remains aligned. \\* \hline
Direct VLM & 0.75 & Object scale is moderately larger than intended, while lid configuration remains aligned. \\* \hline
Claude Code with simulator & 0.50 & Object is substantially oversized relative to the robot workspace. \\* \hline
\methodname{} (Ours), without semantics & 1.00 & Correct scale and drawer configuration. \\ \hline

\multicolumn{3}{|c|}{\textit{Laptop}} \\* \hline
\methodname{} (Ours) & 1.00 & Correct scale and slightly open configuration. \\* \hline
VLM-IVW & 1.00 & Correct scale and slightly open configuration. \\* \hline
Direct VLM & 1.00 & Correct scale and slightly open configuration. \\* \hline
Claude Code with simulator & 0.50 & Object is substantially oversized relative to the robot workspace. \\* \hline
\methodname{} (Ours), without semantics & 1.00 & Correct scale and slightly open configuration. \\ \hline

\multicolumn{3}{|c|}{\textit{Microwave}} \\* \hline
\methodname{} (Ours) & 1.00 & Correct scale and halfway-open door configuration. \\* \hline
VLM-IVW & 1.00 & Correct scale and halfway-open door configuration. \\* \hline
Direct VLM & 1.00 & Correct scale and halfway-open door configuration. \\* \hline
Claude Code with simulator & 0.50 & Object is substantially oversized relative to the robot workspace. \\* \hline
\methodname{} (Ours), without semantics & 1.00 & Correct scale and halfway-open door configuration. \\ \hline

\multicolumn{3}{|c|}{\textit{Stapler}} \\* \hline
\methodname{} (Ours) & 1.00 & Correct scale and slightly open stapler configuration. \\* \hline
VLM-IVW & 0.90 & Stapler body hangs mid-air. \\* \hline
Direct VLM & 0.50 & Joint configuration contains conflicts, leading to an implausible articulated state. \\* \hline
Claude Code with simulator & 0.50 & Object is substantially oversized relative to the robot workspace. \\* \hline
\methodname{} (Ours), without semantics & 0.75 & Scale is correct, but the joint configuration is unrealistic. \\ \hline

\multicolumn{3}{|c|}{\textit{Suitcase}} \\* \hline
\methodname{} (Ours) & 1.00 & Correct scale and slightly open lid configuration. \\* \hline
VLM-IVW & 1.00 & Correct scale and slightly open lid configuration. \\* \hline
Direct VLM & 1.00 & Correct scale and slightly open lid configuration. \\* \hline
Claude Code with simulator & 0.50 & Object is substantially oversized relative to the robot workspace. \\* \hline
\methodname{} (Ours), without semantics & 1.00 & Correct scale and slightly open lid configuration. \\ \hline

\multicolumn{3}{|c|}{\textit{Toilet}} \\* \hline
\methodname{} (Ours) & 1.00 & Correct scale and fully open lid configuration. \\* \hline
VLM-IVW & 0.75 & Lid does not reach the requested fully open configuration. \\* \hline
Direct VLM & 1.00 & Correct scale and fully open lid configuration. \\* \hline
Claude Code with simulator & 0.50 & Toilet is substantially oversized relative to the requested baby-toilet scale. \\* \hline
\methodname{} (Ours), without semantics & 1.00 & Correct scale and fully open lid configuration. \\ \hline

\multicolumn{3}{|c|}{\textit{Trashcan}} \\* \hline
\methodname{} (Ours) & 0.75 & Trashcan scale is moderately larger than requested, while lid state remains partially aligned. \\* \hline
VLM-IVW & 0.25 & Trashcan is too large and the lid is closed despite the prompt requesting a slightly open lid. This is caused by the refined asset's joint limit conflicts. \\* \hline
Direct VLM & 0.75 & Trashcan scale is moderately larger than requested, while lid state remains aligned. \\* \hline
Claude Code with simulator & 0.50 & Trashcan is substantially oversized relative to the requested personal trashcan scale. \\* \hline
\methodname{} (Ours), without semantics & 0.50 & Trashcan lid is closed despite the prompt requesting a slightly open lid. \\ \hline

\multicolumn{3}{|c|}{\textit{Mean Prompt Alignment}} \\ \hline
\methodname{} (Ours) & \textbf{0.97} & Highest mean alignment across evaluated objects. \\ \hline
VLM-IVW & 0.82 & Errors primarily arise from scale misestimation and occasional joint-state mismatch. \\ \hline
Direct VLM & 0.86 & Errors primarily arise from scale misestimation and joint conflicts in articulated mechanisms. \\ \hline
Claude Code with simulator & 0.53 & Errors primarily arise from systematic scale overestimation. \\ \hline
\methodname{} (Ours), without semantics & 0.89 & Errors primarily arise from reduced semantic fidelity in joint configuration, leading to identifying wrong joints. \\ \hline

\end{longtable}
\end{center}

\chapter{\MakeUppercase{Vision-Language-Action Model Evaluation Tasks and Prompt Specifications}}
\label{app:vla-prompts}

This appendix provides the complete specification of behavioral fidelity evaluation tasks and their corresponding language prompts used throughout Chapter~\ref{ch:experiments}.
The evaluation consists of $11$ tasks spanning grasping and articulation-based interactions across multiple object categories. Each task is paired with a natural language instruction that is provided to the policy at inference time.
\begin{table}[h]
\begin{center}
\small
\renewcommand{\arraystretch}{1.2}
\setlength{\tabcolsep}{4pt}

\begin{tabular}{
|>{\centering\arraybackslash}p{0.6cm}
|p{6cm}
|p{7.5cm}|
}
\hline
\textbf{\#} & \textbf{Behavioral Fidelity (VLA) Tasks} & \textbf{Language Prompt} \\
\hline

1 & Stapler Pressing & Press down on the stapler. \\ \hline
2 & Stapler Grasping & Pick up the stapler and put into the bowl. \\ \hline
3 & Box Closing & Close the box lid. \\ \hline
4 & Box Grasping & Pick up the box and put into the bowl. \\ \hline
5 & Toilet Closing & Close the toilet lid. \\ \hline
6 & Microwave Closing & Close the microwave door. \\ \hline
7 & Laptop Folding & Close the laptop. \\ \hline
8 & Bag Grasping & Pick up the bag and put into the bowl. \\ \hline
9 & Suitcase Closing & Close the suitcase lid. \\ \hline
10 & Trashcan Closing & Close the trashcan lid. \\ \hline
11 & Cabinet Closing & Close the drawer of the cabinet. \\ \hline

\end{tabular}

\caption{Behavioral fidelity evaluation tasks and corresponding language prompts. Each task defines a target interaction behavior used to evaluate zero-shot VLA policy performance in both simulation and real-world settings. These prompts are used consistently across all methods.}
\label{tab:task-prompts}

\end{center}
\end{table}

\chapter{\MakeUppercase{Per-Task Vision-
Language-Action Model Evaluation Breakdown}}
\label{app:vla-breakdown}

This appendix provides a detailed per-task breakdown of simulation success rates for each Vision-Language-Action (VLA) policy. Results correspond to the evaluations described in Section~\ref{sec:vla-eval}, with success rates computed over 5 trials per task. Each table reports performance across all methods, along with real-world success rates used for SRCC computation.

\begin{table*}[ht]
\begin{center}
\small
\renewcommand{\arraystretch}{1.2}
\setlength{\tabcolsep}{3pt}

\begin{tabular}{
|p{2.9cm}
|>{\centering\arraybackslash}p{3.5cm}
|>{\centering\arraybackslash}p{2.0cm}
|>{\centering\arraybackslash}p{2.3cm}
|>{\centering\arraybackslash}p{3.2cm}
|>{\centering\arraybackslash}p{1.0cm}|
}
\hline
\textbf{Task} &
\textbf{Human (GenSim2)} &
\textbf{VLM-IVW} &
\textbf{Direct VLM} &
\textbf{\methodname{} (Ours)} &
\textbf{Real} \\
\hline

Stapler Pressing & 0 & 20 & 0 & 0 & 20 \\ \hline
Stapler Grasping & 80 & 60 & 0 & 100 & 100 \\ \hline
Box Closing & 0 & 0 & 0 & 0 & 60 \\ \hline
Box Grasping & 0 & 0 & 0 & 0 & 100 \\ \hline
Toilet Closing & 0 & 60 & 0 & 80 & 80 \\ \hline
Microwave Closing & 0 & 0 & 0 & 40 & 80 \\ \hline
Laptop Folding & 0 & 20 & 0 & 40 & 60 \\ \hline
Bag Grasping & 0 & 0 & 0 & 40 & 100 \\ \hline
Suitcase Closing & 0 & 0 & 0 & 20 & 40 \\ \hline
Trashcan Closing & 0 & 0 & 0 & 20 & 40 \\ \hline
Cabinet Closing & 0 & 20 & 0 & 20 & 60 \\ \hline

\hline
\textbf{Average (\%)} & 7.27 & 16.36 & 0.00 & \textbf{32.73} & 67.27 \\ \hline

\end{tabular}

\caption{Per-task simulation success rates (\%) for $\pi_{0}$.}
\label{tab:vla-pi0}
\end{center}
\end{table*}

\begin{table*}[ht]
\begin{center}
\small
\renewcommand{\arraystretch}{1.2}
\setlength{\tabcolsep}{3pt}

\begin{tabular}{
|p{2.9cm}
|>{\centering\arraybackslash}p{3.5cm}
|>{\centering\arraybackslash}p{2.0cm}
|>{\centering\arraybackslash}p{2.3cm}
|>{\centering\arraybackslash}p{3.2cm}
|>{\centering\arraybackslash}p{1.0cm}|
}
\hline
\textbf{Task} &
\textbf{Human (GenSim2)} &
\textbf{VLM-IVW} &
\textbf{Direct VLM} &
\textbf{\methodname{} (Ours)} &
\textbf{Real} \\
\hline

Stapler Pressing & 0 & 60 & 0 & 40 & 40 \\ \hline
Stapler Grasping & 100 & 100 & 0 & 100 & 100 \\ \hline
Box Closing & 0 & 0 & 0 & 0 & 0 \\ \hline
Box Grasping & 0 & 20 & 0 & 40 & 100 \\ \hline
Toilet Closing & 0 & 0 & 0 & 20 & 60 \\ \hline
Microwave Closing & 0 & 0 & 0 & 40 & 100 \\ \hline
Laptop Folding & 0 & 40 & 0 & 40 & 80 \\ \hline
Bag Grasping & 0 & 0 & 0 & 80 & 80 \\ \hline
Suitcase Closing & 0 & 0 & 0 & 100 & 80 \\ \hline
Trashcan Closing & 0 & 0 & 0 & 40 & 80 \\ \hline
Cabinet Closing & 0 & 0 & 0 & 0 & 20 \\ \hline

\hline
\textbf{Average (\%)} & 9.09 & 20.00 & 0.00 & \textbf{45.45} & 67.27 \\ \hline

\end{tabular}

\caption{Per-task simulation success rates (\%) for $\pi_{0.5}$.}
\label{tab:vla-pi05}
\end{center}
\end{table*}

\begin{table*}[ht]
\begin{center}
\small
\renewcommand{\arraystretch}{1.2}
\setlength{\tabcolsep}{3pt}

\begin{tabular}{
|p{2.9cm}
|>{\centering\arraybackslash}p{3.5cm}
|>{\centering\arraybackslash}p{2.0cm}
|>{\centering\arraybackslash}p{2.3cm}
|>{\centering\arraybackslash}p{3.2cm}
|>{\centering\arraybackslash}p{1.0cm}|
}
\hline
\textbf{Task} &
\textbf{Human (GenSim2)} &
\textbf{VLM-IVW} &
\textbf{Direct VLM} &
\textbf{\methodname{} (Ours)} &
\textbf{Real} \\
\hline

Stapler Pressing & 0 & 40 & 0 & 0 & 0 \\ \hline
Stapler Grasping & 20 & 60 & 0 & 0 & 40 \\ \hline
Box Closing & 0 & 0 & 0 & 40 & 80 \\ \hline
Box Grasping & 0 & 20 & 0 & 0 & 80 \\ \hline
Toilet Closing & 0 & 0 & 0 & 0 & 0 \\ \hline
Microwave Closing & 0 & 0 & 0 & 0 & 0 \\ \hline
Laptop Folding & 0 & 0 & 0 & 20 & 0 \\ \hline
Bag Grasping & 0 & 0 & 20 & 0 & 40 \\ \hline
Suitcase Closing & 0 & 0 & 0 & 100 & 40 \\ \hline
Trashcan Closing & 0 & 0 & 0 & 0 & 20 \\ \hline
Cabinet Closing & 0 & 0 & 0 & 60 & 60 \\ \hline

\hline
\textbf{Average (\%)} & 1.82 & 10.91 & 1.82 & \textbf{20.00} & 32.73 \\ \hline

\end{tabular}

\caption{Per-task simulation success rates (\%) for GR00T N1.6.}
\label{tab:vla-gr00t}
\end{center}
\end{table*}

The per-task results across all three policies highlight consistent trends observed in the main evaluation. Performance varies significantly across tasks and policies, but \methodname{} consistently achieves higher success rates compared to baseline methods.

\chapter{\MakeUppercase{VLM-as-a-Judge Prompting and Evaluation Examples}}
\label{app:vlm-prompt-examples}

This appendix provides representative examples of the VLM-as-a-judge evaluation in Section~\ref{sec:realism-eval}. We include (i) a detailed example illustrating how candidate assets are evaluated based on physical plausibility, and (ii) the full set of task definitions with their VLM--judge language prompts. 

\begin{table*}[ht]
\begin{center}
\small
\renewcommand{\arraystretch}{1.2}
\setlength{\tabcolsep}{3pt}

\begin{tabular}{
|>{\centering\arraybackslash}p{0.4cm}
|p{2.0cm}
|p{5.2cm}
|p{2.8cm}
|p{3.6cm}
|>{\centering\arraybackslash}p{0.9cm}|
}
\hline
\textbf{\#} & \textbf{Method} & \textbf{Description} & \textbf{Pros} & \textbf{Cons} & \textbf{Score} \\
\hline

0 & Real-World &
A real-world video of a robotic arm closing a laptop. The arm moves precisely to the top edge of the screen and applies a steady downward force until the lid is fully closed. &
Perfectly realistic motion; natural timing; correct application of force and gravity; no clipping or jitter. &
None. &
5.0 \\ \hline

1 & Human (GenSim2) &
A 3D animation where the robotic arm moves toward the laptop. The lid begins to close well before the arm makes contact, indicating a lack of physical interaction. &
The laptop lid eventually closes. &
Major violation of cause-and-effect; the lid moves telekinetically without being touched; the arm's movement is not synchronized with the lid's rotation. &
1.5 \\ \hline

2 & VLM-IVW &
The robotic arm moves toward the laptop but stops short of touching it. The laptop remains open throughout the video. &
No clipping occurs. &
Fails to perform the action described in the prompt; the laptop remains static. &
1.0 \\ \hline

3 & Direct VLM &
The robot arm reaches for the laptop, and the lid closes as the arm approaches. While the timing is slightly better than candidate 1, the lid still begins moving before actual contact. &
The task is completed and the arm retracts without clipping. &
The lid moves before contact is established; the rotation of the lid feels mechanical and lacks the weight seen in the ground truth. &
2.0 \\ \hline

4 & \methodname{} (Ours)&
The robot arm moves down and pushes the laptop lid closed. The timing of the lid's movement is the most realistic among the generated candidates, but the arm continues to move downward through the laptop and table after closure. &
The lid closure is correctly triggered by the robot's contact; the initial push feels physically plausible. &
Severe clipping occurs as the arm passes through the laptop and the table; the lid closes with an unnaturally fast, snapping motion. &
3.0 \\ \hline

\end{tabular}

\caption{Example VLM-as-a-judge evaluation output for the Laptop Closing task. Scores are assigned on a 0--5 scale based on consistency of contact, causality, and articulation dynamics, following the protocol described in Section~\ref{sec:vlm-judge}.}
\label{tab:vlm_judge_example}

\end{center}
\end{table*}

\begin{table}[ht]
\begin{center}
\small
\renewcommand{\arraystretch}{1.2}
\setlength{\tabcolsep}{4pt}

\begin{tabular}{
|>{\centering\arraybackslash}p{0.8cm}
|p{3.5cm}
|p{10cm}|
}
\hline
\textbf{\#} & \textbf{Task} & \textbf{Prompt} \\
\hline

1 & Stapler Pressing & The robot tries to press the stapler on the table. \\ \hline
2 & Stapler Grasping  & The robot tries to pick up the stapler and put it into the bowl. \\ \hline
3 & Box Closing      & The robot tries to close the box. \\ \hline
4 & Box Grasping    & The robot tries to pick up the box and put it into the bowl. \\ \hline
5 & Toilet Closing   & The robot tries to close the toilet lid. \\ \hline
6 & Microwave Closing & The robot tries to close the microwave door. \\ \hline
7 & Laptop Closing   & The robot tries to close the laptop. \\ \hline
8 & Bag Grasping      & The robot tries to pick up the bag and put it into the bowl. \\ \hline
9 & Suitcase Closing      & The robot tries to close the suitcase. \\ \hline
10 & Trashcan Closing      & The robot tries to close the trashcan lid. \\ \hline
11 & Cabinet Closing      & The robot tries to close the drawer of the cabinet. \\ \hline
\end{tabular}

\caption{Task definitions and corresponding language prompts used for VLM--as-a-judge evaluation. Each task specifies a target interaction behavior, which is used to guide VLM-as-a-judge evaluation of motion realism.}
\label{tab:task-vlm-prompts}

\end{center}
\end{table}

Together, the appendix shows how realism scores are assigned from observable interaction dynamics and how task prompts define the intended behaviors used for consistent evaluation across tasks.

\chapter{\MakeUppercase{VLM-as-a-Judge Realism Score Breakdown}}
\label{app:vlm-judge-breakdown}

This appendix provides a detailed per-task breakdown of realism scores obtained using Vision-Language Models as judges. We report results for two independent models, Gemini 3.0 and Qwen3-VL 32B. For each task--method pair, scores are reported as mean $\pm$ standard deviation over 10 independent trials. The final column reports the average score across all 11 tasks.

\begin{table*}[ht]
\begin{center}
\small
\renewcommand{\arraystretch}{1.2}
\setlength{\tabcolsep}{4pt}

\begin{tabular}{
|p{3.1cm}
|>{\centering\arraybackslash}p{3.2cm}
|>{\centering\arraybackslash}p{3.4cm}
|>{\centering\arraybackslash}p{2.6cm}
|>{\centering\arraybackslash}p{2.6cm}|
}
\hline
\textbf{Task} &
\textbf{\methodname{} (Ours)} &
\textbf{Human (GenSim2)} &
\textbf{VLM-IVW} &
\textbf{Direct VLM} \\
\hline

\multicolumn{5}{|c|}{\textit{Gemini 3.0}} \\ \hline

Bag Grasping & \textbf{2.35$\pm$0.92} & 1.85$\pm$0.82 & 1.55$\pm$0.99 & 1.60$\pm$0.97 \\ \hline
Box Grasping & \textbf{3.88$\pm$0.54} & 2.25$\pm$0.78 & 2.35$\pm$0.55 & 1.92$\pm$0.55 \\ \hline
Box Closing & 1.20$\pm$0.40 & 1.70$\pm$1.00 & 1.40$\pm$0.49 & \textbf{2.37$\pm$1.16} \\ \hline
Cabinet Closing & \textbf{1.89$\pm$0.80} & 1.70$\pm$0.70 & 1.23$\pm$0.73 & 1.39$\pm$0.72 \\ \hline
Laptop Closing & \textbf{4.10$\pm$0.44} & 3.62$\pm$0.55 & 3.24$\pm$0.96 & 3.44$\pm$0.67 \\ \hline
Microwave Closing & \textbf{1.65$\pm$0.78} & 0.57$\pm$0.23 & 1.27$\pm$0.64 & 1.45$\pm$0.79 \\ \hline
Stapler Grasping & \textbf{3.05$\pm$0.69} & 2.57$\pm$0.80 & 2.04$\pm$0.60 & 1.40$\pm$0.84 \\ \hline
Stapler Pressing & \textbf{2.40$\pm$0.49} & 1.90$\pm$0.42 & 2.19$\pm$0.54 & 2.19$\pm$0.42 \\ \hline
Suitcase Closing & 1.70$\pm$0.51 & 0.70$\pm$0.24 & 1.70$\pm$0.46 & \textbf{1.92$\pm$0.90} \\ \hline
Toilet Closing & \textbf{2.65$\pm$0.84} & 1.25$\pm$0.46 & 1.15$\pm$0.39 & 1.66$\pm$1.05 \\ \hline
Trashcan Closing & 2.60$\pm$0.89 & 0.75$\pm$0.34 & 0.00$\pm$0.00 & \textbf{2.69$\pm$0.73} \\ \hline

\multicolumn{5}{|c|}{\textit{Qwen3-VL 32B}} \\ \hline

Bag Grasping & \textbf{1.90$\pm$0.20} & 1.00$\pm$0.00 & 0.75$\pm$0.34 & 0.75$\pm$0.34 \\ \hline
Box Grasping & \textbf{4.15$\pm$0.32} & 1.80$\pm$0.40 & 3.00$\pm$0.45 & 2.15$\pm$0.71 \\ \hline
Box Closing & 1.90$\pm$0.49 & 1.95$\pm$0.42 & 1.50$\pm$0.50 & \textbf{2.20$\pm$0.71} \\ \hline
Cabinet Closing & 1.94$\pm$0.16 & 2.22$\pm$0.71 & 2.83$\pm$1.15 & \textbf{4.14$\pm$0.43} \\ \hline
Laptop Closing & \textbf{4.40$\pm$0.20} & 1.60$\pm$0.49 & 4.00$\pm$0.32 & 1.05$\pm$0.27 \\ \hline
Microwave Closing & \textbf{4.05$\pm$0.15} & 1.10$\pm$0.30 & 2.10$\pm$0.30 & 1.95$\pm$0.35 \\ \hline
Stapler Grasping & 2.30$\pm$0.46 & 1.70$\pm$0.64 & \textbf{3.10$\pm$0.66} & 0.65$\pm$0.32 \\ \hline
Stapler Pressing & 2.60$\pm$0.44 & 1.40$\pm$0.62 & \textbf{2.90$\pm$0.54} & \textbf{2.90$\pm$0.66} \\ \hline
Suitcase Closing & \textbf{4.05$\pm$0.42} & 1.60$\pm$0.66 & 3.95$\pm$0.42 & 2.10$\pm$0.30 \\ \hline
Toilet Closing & \textbf{2.60$\pm$0.54} & 0.90$\pm$0.30 & \textbf{2.60$\pm$0.83} & 1.45$\pm$1.01 \\ \hline
Trashcan Closing & \textbf{2.65$\pm$0.90} & 0.00$\pm$0.00 & 0.00$\pm$0.00 & 2.35$\pm$0.67 \\ \hline

\end{tabular}

\caption{Per-task realism scores using VLM-as-a-judge evaluation. Rows correspond to individual tasks, and columns correspond to methods. Scores are reported as mean $\pm$ standard deviation over 10 trials on a 0--5 scale. The highest score for each task is bolded.}
\label{tab:vlm_judge_breakdown}
\end{center}
\end{table*}

\chapter{\MakeUppercase{Human Interaction Realism Score Breakdown}}
\label{app:human-eval-breakdown}

This appendix provides a detailed per-object breakdown of human interaction realism scores. Each score reflects expert evaluation of interaction plausibility following the protocol described in Section~\ref{sec:human-telelop}, resulting in the reported Human Realism Score in Section~\ref{sec:realism-eval}. 

\begin{center}
\small
\renewcommand{\arraystretch}{1.2}
\setlength{\tabcolsep}{4pt}

\begin{longtable}{
|p{2.9cm}
|>{\centering\arraybackslash}p{2.1cm}
|>{\centering\arraybackslash}p{2.1cm}
|>{\centering\arraybackslash}p{2.1cm}
|>{\centering\arraybackslash}p{2.1cm}
|>{\centering\arraybackslash}p{2.1cm}
|>{\centering\arraybackslash}p{0.9cm}|
}

\caption{Per-object, per-evaluator human realism scores. Each object is evaluated by 5 independent human raters (evaluators) on a 0--10 scale. The mean across evaluators is reported in the final column; the highest mean score for each object is bolded.}
\label{tab:human_eval_breakdown} \\

\hline
\textbf{Method} & \textbf{Evaluator 1} & \textbf{Evaluator 2} & \textbf{Evaluator 3} & \textbf{Evaluator 4} & \textbf{Evaluator 5} & \textbf{Mean} \\
\hline
\endfirsthead

\hline
\textbf{Method} & \textbf{Evaluator 1} & \textbf{Evaluator 2} & \textbf{Evaluator 3} & \textbf{Evaluator 4} & \textbf{Evaluator 5} & \textbf{Mean} \\
\hline
\endhead

\hline
\multicolumn{7}{r}{\textit{Continued on next page}} \\
\endfoot

\hline
\endlastfoot

\multicolumn{7}{|c|}{\textit{Cabinet}} \\* \hline
\methodname{} (Ours) & 9 & 10 & 8 & 10 & 6 & \textbf{8.6} \\* \hline
Human (GenSim2) & 4 & 4 & 5 & 0 & 4 & 3.4 \\* \hline
VLM-IVW & 4 & 3 & 6 & 2 & 4 & 3.8 \\* \hline
Direct VLM & 3 & 3 & 7 & 0 & 4 & 3.4 \\ \hline

\multicolumn{7}{|c|}{\textit{Microwave}} \\* \hline
\methodname{} (Ours) & 9 & 8 & 8 & 10 & 9 & \textbf{8.8} \\* \hline
Human (GenSim2) & 1 & 1 & 3 & 0 & 3 & 1.6 \\* \hline
VLM-IVW & 2 & 1 & 4 & 2 & 6 & 3.0 \\* \hline
Direct VLM & 2 & 1 & 4 & 2 & 6 & 3.0 \\ \hline

\multicolumn{7}{|c|}{\textit{Laptop}} \\* \hline
\methodname{} (Ours) & 7 & 9 & 8 & 6 & 7 & \textbf{7.4} \\* \hline
Human (GenSim2) & 3 & 1 & 4 & 2 & 3 & 2.6 \\* \hline
VLM-IVW & 5 & 9 & 8 & 7 & 7 & 7.2 \\* \hline
Direct VLM & 3 & 1 & 4 & 3 & 3 & 2.8 \\ \hline

\multicolumn{7}{|c|}{\textit{Stapler}} \\* \hline
\methodname{} (Ours) & 3 & 6 & 3 & 5 & 4 & 4.2 \\* \hline
Human (GenSim2) & 2 & 1 & 3 & 4 & 2 & 2.4 \\* \hline
VLM-IVW & 2 & 9 & 5 & 7 & 6 & \textbf{5.8} \\* \hline
Direct VLM & 1 & 1 & 2 & 0 & 0 & 0.8 \\ \hline

\multicolumn{7}{|c|}{\textit{Trashcan}} \\* \hline
\methodname{} (Ours) & 7 & 8 & 8 & 10 & 7 & \textbf{8.0} \\* \hline
Human (GenSim2) & 1 & 0 & 4 & 0 & 0 & 1.0 \\* \hline
VLM-IVW & 0 & 0 & 0 & 0 & 0 & 0.0 \\* \hline
Direct VLM & 1 & 3 & 5 & 1 & 4 & 2.8 \\ \hline

\multicolumn{7}{|c|}{\textit{Toilet}} \\* \hline
\methodname{} (Ours) & 8 & 10 & 8 & 8 & 8 & \textbf{8.4} \\* \hline
Human (GenSim2) & 1 & 2 & 4 & 0 & 3 & 2.0 \\* \hline
VLM-IVW & 8 & 9 & 5 & 10 & 9 & 8.2 \\* \hline
Direct VLM & 2 & 2 & 4 & 1 & 1 & 2.0 \\ \hline

\multicolumn{7}{|c|}{\textit{Bag}} \\* \hline
\methodname{} (Ours) & 2 & 6 & 8 & 7 & 3 & \textbf{5.2} \\* \hline
Human (GenSim2) & 2 & 1 & 3 & 5 & 2 & 2.6 \\* \hline
VLM-IVW & 2 & 4 & 7 & 5 & 4 & 4.4 \\* \hline
Direct VLM & 2 & 3 & 7 & 4 & 4 & 4.0 \\ \hline

\multicolumn{7}{|c|}{\textit{Suitcase}} \\* \hline
\methodname{} (Ours) & 4 & 2 & 6 & 4 & 9 & \textbf{5.0} \\* \hline
Human (GenSim2) & 3 & 1 & 3 & 0 & 1 & 1.6 \\* \hline
VLM-IVW & 1 & 0 & 2 & 0 & 0 & 0.6 \\* \hline
Direct VLM & 3 & 1 & 4 & 3 & 5 & 3.2 \\ \hline

\multicolumn{7}{|c|}{\textit{Box}} \\* \hline
\methodname{} (Ours) & 7 & 7 & 9 & 8 & 8 & \textbf{7.8} \\* \hline
Human (GenSim2) & 2 & 3 & 4 & 1 & 2 & 2.4 \\* \hline
VLM-IVW & 2 & 3 & 4 & 0 & 5 & 2.8 \\* \hline
Direct VLM & 2 & 3 & 5 & 2 & 5 & 3.4 \\ \hline

\end{longtable}
\end{center}

\chapter{\MakeUppercase{Large-Scale Object Set for Refinement}}
\label{app:large-scale-objects}

This appendix lists all 93 assets spanning 10 different object categories used in the large-scale evaluation described in Section~\ref{sec:scaling-eval}. Each object is sourced from PartNet-Mobility and paired with a natural language prompt used for physical refinement.

\begin{table*}[ht]
\begin{center}
\small
\renewcommand{\arraystretch}{1.2}
\setlength{\tabcolsep}{4pt}

\begin{tabular}{
|p{2.1cm}
|p{6.5cm}
|p{7.0cm}|
}
\hline
\multicolumn{1}{|c|}{\textbf{Object}} &
\multicolumn{1}{c|}{\textbf{PartNet-Mobility IDs}} &
\multicolumn{1}{c|}{\textbf{Prompt}} \\
\hline
Box & 47645, 48492, 100129, 100141, 100154, 100162, 100174, 100189, 100191, 100194 & A small box with slightly open rotational lid. Low damping for a smooth hinge behavior. \\ \hline

Folding Chair & 100520, 100521, 100523, 100526, 100531, 100532, 100557, 100561, 100562, 100568 & I want a very small folding chair for a baby that is easy to fold and unfold, with its seat open all the way in normal position. \\ \hline

Keyboard & 7619, 12727, 12738, 12829, 12834, 12836, 12838, 12851, 12880 & A standard plastic computer keyboard with light, spring-like keys for easy pressing.  \\ \hline

Knife & 101052, 101054, 101057, 101059, 101062, 101068, 101079, 101080, 101085, 101095 & I want a regular size folding knife that fits in hand. Low damping for hinge-like behavior. \\ \hline

Laptop & 9748, 9912, 9960, 9968, 9992, 9996, 10040 & I want a slightly open laptop. Low damping for smooth operation. \\ \hline

Microwave & 7119, 7128, 7167, 7236, 7263, 7265, 7273, 7296 & I want a microwave opening its door halfway. Low damping for hinge-like behavior.\\ \hline

Scissors & 10449, 10450, 10495, 10499, 10502, 10537, 10557, 10558, 10559 & A wide open pair of scissors with two intersecting blades connected through a rotational hinge joint, low damping for easy opening and closing motion. \\ \hline

Stapler & 102990, 103095, 103099, 103100, 103104, 103111, 103113, 103271, 103273, 103275 & I want a slightly open stapler, with spring behavior in its joint for pressing/stapling. \\ \hline

Toilet & 101319, 101320, 101323, 102619, 102620, 102621, 102622, 102625, 102628, 102629 & I want a very small toilet for a baby, with its lid open all the way. Low damping for hinge-like behavior. \\ \hline

Trashcan & 4108, 10357, 10584, 11124, 11229, 11259, 11279, 11361, 11818, 11951 & A very small, personal trash can with its lid slightly open. The lid hinge has low damping and holds its position. \\ \hline

\end{tabular}

\caption{All 93 assets across 10 object categories used in the large-scale refinement experiment.}
\label{tab:large-scale-objects}

\end{center}
\end{table*}

\chapter{\MakeUppercase{Author Contributions}}
\label{app:personal-contributions}

This thesis is solely authored by the candidate and is based in part on collaborative work that has resulted in a co-authored publication currently under review. The contributions of each author, as they relate to the material presented in this thesis, are detailed below:

\textbf{Anh Quan Pham}:
\begin{itemize}
\item Led the formulation of the evaluation methods described in Sections~\ref{sec:3.1intro}, \ref{sec:alignment}, \ref{sec:RL-theory}, and \ref{sec:human-telelop}.
\item Led the implementation and experimental evaluation for all baselines and experiments reported in Sections~\ref{sec:exp-setup}–\ref{sec:rl-eval}, \ref{sec:realism-eval} (human teleoperation experiments), and \ref{sec:scaling-eval}–\ref{sec:efficiency}.
\item Iteratively developed the core architecture proposed in Chapter~\ref{ch:method} to produce the final system, including support for all types of objects and parallel, large-scale refinement.
\item The Introduction (Chapter~\ref{ch:introduction}) and Conclusion (Chapter~\ref{ch:conclusion}) were written largely independently and extend beyond the scope of the co-authored publication.
\end{itemize}

\textbf{Luyang Hu}:
\begin{itemize}
\item Formalized and implemented the early version of the framework proposed in Chapter~\ref{ch:method}.
\item Led the initial formulation of the methods described in Sections~\ref{sec:stability} and \ref{sec:vla-eval-method}.
\item Developed numerical evaluation metrics (e.g., penetration depth, pass rates) reported in Section~\ref{sec:fidelity-eval}, and conducted $\pi_{0.5}$ simulation experiments in Section~\ref{sec:vla-eval}.
\item Implemented the ablation studies described in Subsection~\ref{sec:ablations}.
\item Formalized the method in Section~\ref{sec:vlm-judge} and implemented the corresponding VLM-judge evaluation in Section~\ref{sec:realism-eval}.
\end{itemize}

\textbf{Kaitian Chao}:
\begin{itemize}
\item Conducted real-world robot experiments across all three Vision-Language-Action models. These results are utilized in Sections \ref{sec:vla-eval} and \ref{sec:realism-eval} (VLM-judge experiments).
\end{itemize}

\textbf{Sagnik Anupam}:
\begin{itemize}
\item Proposed baselines and ablations described in Section~\ref{sec:baselinesandablations}.
\item Contributed meaningfully to Chapter~\ref{ch:related_work}: Related Work.
\end{itemize}

\textbf{George Jiayuan Gao, Tianyou Wang, and Junyao Shi:}
\begin{itemize}
    \item Mentorship and conceptual guidance.
\end{itemize}

\chapter{\MakeUppercase{Prompt Template: Physical Property Generation}}
\label{app:prompt_template}
\begin{tcolorbox}[
  colback=gray!5,
  colframe=black,
  title={URDF Overlay Generation System Prompt},
  fonttitle=\bfseries,
  breakable
]
\begin{lstlisting}[basicstyle=\small\ttfamily, columns=fullflexible, breaklines=true]
"""Centralized prompts for URDF refinement VLM interactions.

This module contains all prompts used throughout the urdf_refiner package
for Vision-Language Model (VLM) interactions.
"""

# =============================================================================
# System Prompts
# =============================================================================

URDF_EXPERT_SYSTEM_PROMPT = (
    "You are a robotics expert specializing in URDF physical property specification "
    "with strong visual reasoning capabilities."
)

STATE_REFINER_SYSTEM_PROMPT = (
    "You are a robotics simulation expert. Adjust joint positions to reduce "
    "self-collisions. Output strict JSON only."
)


# =============================================================================
# Overlay Generation Prompt
# =============================================================================

OVERLAY_GENERATION_PROMPT = """
You are a robotics/physics expert tasked with generating physically plausible property edits for a URDF asset and returning them as a **structured JSON overlay**.

This overlay is used for **Isaac Sim / PhysX articulations**. Most assets are **PASSIVE** (no active joint drives/servos), so joint parameters must be chosen accordingly.

You are provided with:
1) Rendered images of the assembled object from multiple viewpoints (Front, Back, Left, Right, Perspective)
2) User guidance describing materials, properties, and expected behavior
3) Technical data including dimensions, mesh volumes, and any existing URDF properties

Your job: combine ALL sources of information to produce a single JSON overlay that makes the object physically realistic and stable in simulation.

=====================================================================
CONFLICT RESOLUTION PRIORITY (must follow)
User Guidance  >  Visual Observation  >  Conservative Defaults
If user guidance conflicts with visuals, obey user and note the discrepancy in validation_notes.
=====================================================================

## CRITICAL: Multi-Modal Reasoning (Vision + Guidance + Data)
YOUR APPROACH:
- START with visual observation (what you see)
- CROSS-REFERENCE with user guidance (what the user tells you)
- VALIDATE with technical data (volumes, dimensions, existing properties)
- RESOLVE conflicts with the priority rule above

---------------------------------------------------------------------
STEP 1: Visual Appearance Analysis (required)
---------------------------------------------------------------------
For EACH visible part/link, observe:
1) Surface texture & finish:
   - shiny/reflective -> likely metal or coated
   - matte/colored -> plastic/painted
   - dark matte/grippy -> rubber/soft polymer
2) Geometry & thickness:
   - thick/solid-looking -> likely heavier / hollow_factor closer to 1
   - thin shell/cover -> likely lighter / hollow_factor lower
3) Functional role:
   - base/support parts usually heavier
   - moving arms/lids usually lighter
Record a short _visual_notes per link.

---------------------------------------------------------------------
STEP 2: Material inference (required)
---------------------------------------------------------------------
Infer material class from visuals unless user overrides:
- METAL -> density 2700--8000 kg/m^3 (Al 2700, Steel 7850, Brass 8500)
- PLASTIC -> density 900--1400 kg/m^3 (PP 900, ABS 1040, PVC 1380)
- RUBBER -> density 1000--1500 kg/m^3 (use ~1200 if uncertain)
If uncertain, pick a conservative mid-range and explicitly note uncertainty.

---------------------------------------------------------------------
STEP 3: Uniform scaling (REQUIRED)
---------------------------------------------------------------------
If target dimensions exist in object_info.scaling_analysis -> use the recommended scale exactly.
Otherwise infer a realistic real-world size for the object category (tabletop/handheld):
- Typical handheld cross-section: 0.03--0.07 m
- Typical length: 0.10--0.20 m for elongated objects
Constraint: at least 2 final dimensions must be <= 0.075 m (gripper opening).
Compute: scale = target_size / current_size, and verify final scaled dimensions are realistic.

In validation_notes include:
"SIZE REASONING: Typical real-world [type] ~= [XxYxZ] m. Current dims [AxBxC] m. Scale s=[...]. Final dims [PxQxR] m."

---------------------------------------------------------------------
STEP 4: Mass calculation (required)
---------------------------------------------------------------------
For each link with volume data:
mass = volume_m^3 * density_kg_m^3 * (uniform_scale_factor^3) * hollow_factor

Choose hollow_factor based on appearance:
- solid-looking: 0.8--1.0
- moderately hollow/shelled: 0.4--0.8
- very thin shell: 0.2--0.5

Then adjust slightly based on role:
- base/static parts: should carry most mass
- moving parts: lighter than base
- tiny decorative parts: 0.001--0.01 kg

In validation_notes, show at least a few explicit mass computations.

---------------------------------------------------------------------
STEP 5: Inertia (required)
---------------------------------------------------------------------
Generate inertia for each link (positive definite), using a simple shape approximation
(box/cylinder) from scaled dimensions or bounding extents. Set off-diagonals to 0 unless
strongly justified. Ensure magnitudes scale consistently with mass and size.

---------------------------------------------------------------------
STEP 6: Joint dynamics (PASSIVE by default, Isaac Sim / PhysX)
---------------------------------------------------------------------
Most objects are NOT driven. Therefore:
- Treat joints as PASSIVE by default (no servo/drive).
- PASSIVE joints MUST always include all three dynamics parameters:
  - damping
  - friction
  - stiffness
- This is mandatory for all joints in the object (not optional).
- If spring behavior is not intended, set stiffness = 0.0 (do NOT omit stiffness).
- Only set stiffness > 0 if user explicitly requests spring-return behavior OR you have
  strong evidence of a spring mechanism AND you intentionally model it as a spring.

### Units (SI) --- IMPORTANT
Revolute/continuous (angular):
- damping:  N*m*s / rad
- friction: N*m
- stiffness (ONLY if modeling a spring/return): N*m / rad

Prismatic (linear):
- damping:  N*s / m
- friction: N
- stiffness (ONLY if modeling a spring/return): N / m

### What each term means (PASSIVE interpretation)
- damping: viscous resistance proportional to joint velocity; dissipates energy; reduces jitter/oscillation
- friction: Coulomb/static-like resistance; helps stop small drift; too high can cause stick-slip
- stiffness: spring restoring torque/force toward a reference; causes spring-back; set to 0.0 unless desired

### Passive damping policy (stability-first)
To reduce oscillation/jitter in simulation, set passive damping slightly higher by default than "minimal physics":

Revolute/continuous PASSIVE defaults:
- damping:  0.05--0.5 N*m*s/rad  (use ~0.15 as a safe default)
- friction: 0.005--0.05 N*m
- stiffness: 0.0 by default (must still be present)

Prismatic PASSIVE defaults:
- damping:  0.5--5.0 N*s/m       (use ~1.5 as a safe default)
- friction: 0.1--2.0 N
- stiffness: 0.0 by default (must still be present)

Escalation rule:
- If the asset shows persistent oscillation or joint jitter in contact scenes, increase damping by 2x--5x,
  and add small friction only if needed. Avoid excessive friction that causes stick-slip.
- Only use very large damping if user requests "super high damping" OR stability demands it and you note it.

### Spring-eligible hinges (special handling)
Some objects include an internal spring or spring-like return mechanism (e.g., staplers, pliers, clothespins, tongs, nail clippers).
You must explicitly check for this possibility using the kinematic structure + visuals + guidance.

Identify a hinge as "spring-eligible" if MOST of the following are true:
- The object is a tool/clamp-like mechanism (pliers/tongs/clip/stapler) OR user guidance mentions "spring/return/snap".
- The kinematic structure has a primary revolute joint connecting two long arms/levers (high leverage geometry).
- The rendered images show a visible coil/leaf spring, or the link naming/semantics indicate "spring", "clip", "handle", "jaw".

If a hinge is spring-eligible:
- Default to PASSIVE unless user explicitly requests spring behavior.
- If user requests spring-return OR visual evidence is strong, model it by setting stiffness > 0 (spring),
  and pair it with damping to avoid oscillations.
- If user does NOT request spring-return and evidence is weak/ambiguous, keep stiffness = 0 and instead
  increase damping/friction slightly for stability; note uncertainty in validation_notes.

### Spring modeling (ONLY when intended)
If spring-return is intended (user requested OR strong visual evidence):
- Revolute spring: stiffness ~ 0.05--2.0 N*m/rad (scale up for larger/heavier arms)
- Revolute damping: 0.01--0.2 N*m*s/rad (enough to prevent ringing)
- Prismatic spring: stiffness ~ 50--500 N/m
- Prismatic damping: 0.2--5.0 N*s/m

Choose larger values for larger/heavier links (greater moving mass/inertia) and for faster return.
In validation_notes, always state:
- Why you believe the hinge is spring-eligible (kinematics/visual/guidance evidence)
- Whether you enabled stiffness (spring) or kept it passive
- The stability tradeoff (damping/friction increases)

### Qualitative mapping (if user uses words)
- "super low" -> near minimum
- "low" -> lower quartile
- "medium/moderate" -> midrange
- "high" -> upper quartile
- "super high" -> near maximum of the chosen range (avoid destabilizing values)

### "Stays where placed" (important nuance)
If user says "stays where placed":
- set stiffness ~= 0 (no spring-back),
- raise damping to medium/high within passive ranges,
- add small friction only if needed to stop drift.

### Absolute output requirements (NEVER violate)
- ALWAYS ALWAYS include damping, friction, and stiffness for every joint in output JSON/URDF edits.
- If stiffness does not need modification, set stiffness = 0.0 (never leave it out).
- For revolute/continuous joints, all output numeric values are in radians-based SI units.
- If any input hint/source uses degrees, convert degrees to radians BEFORE output.
- Never output degree values for angular joints.

---------------------------------------------------------------------
STEP 7: Object state / initial joint positions (ONLY if user specifies)
---------------------------------------------------------------------
Only include "initial_joint_positions" if the user explicitly requests a state
(open/closed/extended/retracted/half-open/etc.).

Process:
1) OBSERVE the renders to determine what joint position 0 visually corresponds to.
2) Compare desired state vs observed state.
3) Use joint limits to choose a value near 0 (same state) or near limit (opposite state).
4) Document reasoning in validation_notes.

Unit handling for state values (critical):
- Revolute/continuous joint positions MUST be output in radians.
- If user guidance, metadata, or internal reasoning references degrees, convert to radians before output.
- Never emit degree values in initial_joint_positions.

Do NOT include initial_joint_positions if the user did not request a state.

---------------------------------------------------------------------
STEP 8: Contact properties (optional)
---------------------------------------------------------------------
Only include global material_properties if needed for stable contact (e.g., rubber pads, very slippery object).
If unsure, omit contact_stiffness/contact_damping and only set friction/restitution conservatively.

=====================================================================
Object Information:
{object_info}
=====================================================================

## Output Format (JSON ONLY)
Return a JSON object exactly matching this structure:

{{
  "global_modifications": {{
    "uniform_scale_factor": 0.1,
    "material_properties": {{
      "friction": 0.5,
      "restitution": 0.1,
      "contact_stiffness": 1000,
      "contact_damping": 10
    }}
  }},
  "link_modifications": {{
    "link_name": {{
      "mass": 0.5,
      "inertia": {{
        "ixx": 0.001, "iyy": 0.001, "izz": 0.001,
        "ixy": 0.0, "ixz": 0.0, "iyz": 0.0
      }},
      "center_of_mass": [0.0, 0.0, 0.0],
      "material_override": {{
        "friction": 0.8
      }},
      "_action": "new",
      "_visual_notes": "What you observed + inferred material + hollow/solid judgment"
    }}
  }},
  "joint_modifications": {{
    "joint_name": {{
      "damping": 0.15,
      "friction": 0.01,
      "stiffness": 0.0,
      "_action": "new"
    }}
  }},
  "initial_joint_positions": {{
    "joint_name": 1.57,
    "_reasoning": "Only if user requested a state; explain limit-based choice"
  }},
  "validation_notes": [
    "SIZE REASONING: ...",
    "STATE CONFIGURATION: ... (only if used)",
    "VISUAL MATERIAL REASONING: ...",
    "MASS CALCULATIONS: ...",
    "INERTIA ASSUMPTIONS: ...",
    "JOINT DYNAMICS: ... (explicitly note PASSIVE assumption; units; stiffness always present and 0 unless spring)",
    "UNIT CHECK: ... (explicitly confirm all angular values are radians; any degree inputs converted)",
    "SPRING CHECK: ... (only for spring-eligible hinges; explain evidence and choice)",
    "USER GUIDANCE OVERRIDES: ...",
    "UNCERTAINTIES/WARNINGS: ..."
  ]
}}

## Critical Requirements
1) Output ONLY JSON (no extra text)
2) uniform_scale_factor MUST be uniform (one scalar)
3) Every link MUST have mass + inertia set
4) Every joint MUST have damping + friction + stiffness set; if spring-return is not intended use stiffness = 0.0
5) Inertia tensors MUST be positive definite
6) Use visual observations explicitly in _visual_notes and validation_notes
7) Total mass must be realistic for the object category and scaled size
8) All revolute/continuous values MUST be radians; convert any degree-based inputs before output
"""


# =============================================================================
# State Refinement VLM Prompt Template
# =============================================================================

STATE_REFINEMENT_PROMPT_TEMPLATE = """You are refining an articulated object's joint positions.

IMPORTANT:
- Physics collision data is authoritative for penetration
- Images are authoritative for visual state match

You are given 4 rendered views: [perspective, front, side, top]

## User's desired state
{user_hint}

## Physics collision score
- penetration_sum_m: {penetration_sum:.6f}
- severity: {severity}
- contact_count: {contact_count}

## Collision context
- description: {collision_description}
- top_penetrating_pairs: {top_pen_sums}
- resolved_pairs: {resolved_pairs}
- direct_colliding_joints: {direct_colliding_joints}

## Focus joints (adjust these)
{focus_joints_list}

Joint info:
{focus_info}

## Output format (STRICT JSON ONLY)

1) STATE CHECK: Does the object visually match the user's desired state?
2) If state_ok is true, do COLLISION REDUCTION

Step size guidance:
- If state_ok is false: use larger deltas (10-40% of joint range)
- If state_ok is true + large penetration (>0.02m): medium deltas (5-25%)
- If penetration is low (<0.01m): small deltas

If penetration_sum_m <= 0.002 AND state_ok is true, approve:
{{"approved": true, "state_ok": true, "action": "approve", "reason": "..."}}

Otherwise, return:
{{
  "approved": false,
  "state_ok": true/false,
  "action": "collision_reduction" or "state_correction",
  "reason": "Explain which pair you're targeting and how",
  "joint_updates": [
    {{"joint": "joint_name", "delta": -0.05}}
  ]
}}

Return ONLY valid JSON. No markdown.
"""


# =============================================================================
# Self-Collision Warning Messages
# =============================================================================

SELF_COLLISION_WARNING = (
    "SELF-COLLISION DETECTED: The object has mesh interpenetration that could not be resolved. "
    "If the visual state looks correct, consider enabling 'Disable Self-Collision' option. "
    "This will skip self-collision checks in Isaac Sim fidelity tests."
)

SELF_COLLISION_WARNING_NO_REFINEMENT = (
    "SELF-COLLISION DETECTED: The object has mesh interpenetration in its default/requested state. "
    "This may cause physics instability in Isaac Sim. If the visual state looks correct, "
    "enable 'Disable Self-Collision' option to skip self-collision checks in fidelity tests."
)


# =============================================================================
# Helper function to build state refinement prompt
# =============================================================================

def build_state_refinement_prompt(
    current_positions: dict,
    target_positions: dict,
    score,
    joint_limits: dict,
    user_hint: str,
    extracted_info: dict,
    collision_analysis: dict,
    focus_joints: list,
    iteration: int,
) -> str:
    """Build the VLM feedback prompt for state refinement.
    
    Args:
        current_positions: Current joint positions
        target_positions: Target joint positions
        score: ContactScore object with penetration info
        joint_limits: Dict mapping joint names to (lower, upper) limits
        user_hint: User's desired state description
        extracted_info: Extracted URDF information
        collision_analysis: Analysis of current collisions
        focus_joints: List of joints to focus on
        iteration: Current iteration number
        
    Returns:
        Formatted prompt string
    """
    import json as json_module
    
    # Severity assessment
    if score.penetration_sum < 0.01:
        severity = "minor"
    elif score.penetration_sum < 0.05:
        severity = "moderate"
    else:
        severity = "significant"
    
    # Extract semantic information
    urdf_joints = extracted_info.get("urdf_structure", {}).get("joints", {}) or {}
    semantics = extracted_info.get("semantics", {})
    
    # Build link->semantic mapping
    link_to_semantic = {}
    if isinstance(semantics, dict):
        for link, info in semantics.items():
            if isinstance(info, dict) and info.get("semantic_name"):
                link_to_semantic[str(link)] = str(info["semantic_name"])
    
    # Build joint->semantic mapping via child links
    joint_semantics = {}
    for joint_name, info in urdf_joints.items():
        if isinstance(info, dict):
            child = info.get("child", "")
            if child and child in link_to_semantic:
                joint_semantics[str(joint_name)] = link_to_semantic[child]
    
    def joint_info(jname: str) -> dict:
        ji = urdf_joints.get(jname, {}) if isinstance(urdf_joints.get(jname), dict) else {}
        child_link = ji.get("child")
        limits = joint_limits.get(jname)
        lo, hi = (float(limits[0]), float(limits[1])) if limits else (None, None)
        cur = float(current_positions.get(jname, 0.0))
        jtype = ji.get("type")
        unit = "radians" if jtype in ("revolute", "continuous") else "meters" if jtype == "prismatic" else "units"
        return {
            "name": jname,
            "type": jtype,
            "child_link": child_link,
            "semantic": joint_semantics.get(jname) or (link_to_semantic.get(child_link) if child_link else None),
            "limit": {"lower": lo, "upper": hi} if limits else None,
            "range": hi - lo if limits and hi > lo else None,
            "unit": unit,
            "current": cur,
            "target": float(target_positions.get(jname, cur)),
        }
    
    focus_info = {j: joint_info(j) for j in focus_joints[:8]}
    top_pairs = collision_analysis.get("resolved_pairs", [])[:6]
    top_pen_sums = score.penetrating_pair_sums[:6]
    
    return STATE_REFINEMENT_PROMPT_TEMPLATE.format(
        user_hint=user_hint or "Not specified",
        penetration_sum=score.penetration_sum,
        severity=severity,
        contact_count=score.contact_count,
        collision_description=collision_analysis.get('collision_description', ''),
        top_pen_sums=json_module.dumps(top_pen_sums, indent=2),
        resolved_pairs=json_module.dumps(top_pairs, indent=2),
        direct_colliding_joints=json_module.dumps(collision_analysis.get('direct_colliding_joints', [])),
        focus_joints_list=json_module.dumps(list(focus_info.keys())),
        focus_info=json_module.dumps(focus_info, indent=2),
    )

\end{lstlisting}
\end{tcolorbox}

\end{append}

\begin{bibliof}
\bibliography{bibliography}
\end{bibliof}
\end{document}